\documentclass{article}

\usepackage{microtype}
\usepackage{graphicx}
\usepackage{booktabs} % for professional tables

\usepackage{hyperref}

\usepackage[accepted]{icml2025}
\makeatletter
\renewcommand{\ICML@appearing}{\textit{Preprint.}}
\makeatother

\usepackage{amsmath}
\usepackage{amssymb}
\usepackage{mathtools}
\usepackage{amsthm}

\usepackage[capitalize,noabbrev]{cleveref}
\usepackage{tcolorbox}

\theoremstyle{plain}
\newtheorem{theorem}{Theorem}[section]

\theoremstyle{definition}
\newtheorem{definition}[theorem]{Definition}

\theoremstyle{remark}

\usepackage[disable,textsize=tiny]{todonotes}

\usepackage[utf8]{inputenc} % allow utf-8 input
\usepackage[T1]{fontenc}    % use 8-bit T1 fonts
\usepackage{url}            % simple URL typesetting
\usepackage{booktabs}       % professional-quality tables
\usepackage{amsfonts}       % blackboard math symbols
\usepackage{nicefrac}       % compact symbols for 1/2, etc.
\usepackage{microtype}      % microtypography
\usepackage{xcolor}         % colors

 \usepackage{amsmath}
\usepackage{graphicx}
\usepackage{enumitem}

\usepackage{booktabs}
\usepackage{array}     % for p{} column types
\usepackage{wrapfig}
\usepackage{subcaption}

\newcommand{\yue}[1]{}
\newcommand{\ZZ}[1]{}

\newcommand{\model}{\textbf{GCB}}

\icmltitlerunning{Exploring Concept Subspace for Self-explainable Text-Attributed Graph Learning}

\begin{document}

\twocolumn[
\icmltitle{Exploring Concept Subspace for Self-explainable Text-Attributed \\Graph Learning}
% Exploring Concept Subspace for Self-explainable Learning on Text-attributed Graphs?


\icmlsetsymbol{equal}{*}

\begin{icmlauthorlist}
\icmlauthor{Xiaoxue Han}{ste}
\icmlauthor{Libo Zhang}{ste}
\icmlauthor{Zining Zhu}{ste}
\icmlauthor{Yue Ning}{ste}
\end{icmlauthorlist}

\icmlaffiliation{ste}{Department of Computer Science, Stevens Institute of Technology, Hoboken, NJ, USA}

\icmlcorrespondingauthor{Xiaoxue Han}{xhan26@stevens.edu}

\icmlkeywords{Graph Neural Networks, Concept Bottleneck, Interpretability, Text-Attributed Graphs, Machine Learning}

\vskip 0.3in
]



\printAffiliationsAndNotice{}  % shows affiliations in preprint

\begin{abstract}
 We introduce Graph Concept Bottleneck (\model{}) as a new paradigm for self-explainable text-attributed graph
learning. \model{}  maps graphs into a subspace—\textit{concept bottleneck}—where each concept is a meaningful phrase, and predictions are made based on the activation of these concepts. Unlike existing interpretable graph learning methods that primarily rely on subgraphs as explanations, the concept bottleneck provides a new form of interpretation. To refine the concept space, we apply the information bottleneck principle to focus on the most relevant concepts. This not only yields more concise and faithful explanations but also explicitly guides the model to ``think'' toward the correct decision. We empirically show that \model{} achieves intrinsic interpretability with accuracy on par with black-box Graph Neural Networks. Moreover, it delivers better performance under distribution shifts and data perturbations, showing improved robustness and generalizability, benefitting from concept-guided prediction.

\end{abstract}

\section{Introduction}
Graph Neural Networks (GNNs)~\cite{kipf2017GCN, veličković2018GAT, yun2020GT, xu2018how} have demonstrated strong performance across a wide range of domains. In many real-world applications, graphs are text-attributed, where node or edge features are expressed in natural language. For example, in social networks, each node may represent a user or a discussion thread and be associated with textual content such as posts or comments; in e-commerce graphs, a product node is often accompanied by its textual description.
Such text-attributed graphs (TAGs)~\cite{Yan2023TAG} are ubiquitous in practice and provide rich semantic information. As text-attributed graph learning is increasingly deployed in high-stakes domains, \emph{trustworthiness} has emerged as a critical concern. 

An effective way to build trust is to provide transparent interpretations of the prediction process~\cite{kakkad2023-survey}.
Intrinsic interpretability, which enables models to explain their predictions directly without relying on post-hoc explanations, becomes a particularly desirable property for GNN-based models~\cite{miao2022-GIB}. 
Most self-explanable GNNs (SE-GNNs)~\cite{azzolin2025beyond, Wang2025prototype, liu2025faithfulclasslevelselfexplainabilitygraph} focus on extracting the most informative yet compressed causal subgraph, which is assumed to be responsible for the prediction and is used for both decision-making and explanation. 
However, while such subgraphs are typically smaller and contain less redundant information, they are still graphs—often complex and difficult to interpret.
% Even with emerging visualization tools~\cite{Bastian_Heymann_Jacomy_2009, shannon2003cytoscape, chimani2014ogdf, gansner2000graphviz, hagberg2008exploring}, 
It remains challenging for humans to understand these explanations, especially in domains where expert knowledge is lacking or the graph structure is intricate.
% Although compressed subgraphs take a step forward in making predictions more interpretable, there is still a substantial gap between these explanations and what humans can readily understand.

We narrow the gap between model predictions and human understanding by introducing an intermediate representation - \emph{concept bottleneck}. Specifically, the input graph is mapped to a concept subspace that captures its activations over a set of semantically meaningful concepts. These activations are then mapped to the label through a few feedforward layers for label prediction. In this way, the concept activations serve a dual purpose: they drive the prediction and simultaneously provide explanations for it.
Although the general workflow is straightforward, adapting it to graph prediction tasks is non-trivial and introduces several challenges: 
(1) \textit{Concept selection:} It is labor-intensive to define concept sets relevant to the prediction task, and graphs often represent abstract structures (e.g., social networks), making it difficult to define and select meaningful, human-interpretable concepts. 
(2) \textit{Concept alignment:} It remains unclear how to effectively map the input domain (graphs) to the concept domain (language). Unlike vision-language tasks—where models like \textsc{CLIP}~\cite{radford2021learningtransferablevisualmodels} provide off-the-shelf alignment—graphs exhibit irregular structures and high variability across domains, and no such readily applicable model exists. Consequently, a concept predictor must be carefully designed to ensure faithful explanations.

 % incorporating the Concept Bottleneck Model (CBM)~\cite{shin2023closerlookCBM, shang2024incrementalCBM, yuksekgonul2023posthocCBM, kim2023probabilisticCBM, koh2020conceptbottleneckmodels, oikarinen2023labelfreeconceptbottleneckmodels}. Having been widely explored in computer vision,  the core idea of CBM is that, rather than predicting the target label \( \hat{y} \) (e.g., bird species) directly from the input image \( x \), the model first predicts a set of high-level concepts \( \hat{c} \) (e.g., “wing color” or “beak shape”), and then uses those concepts to predict the final label. These concepts are predefined in natural language and are directly understandable to humans, enabling interpretation by design. 

% However, generalizing Concept Bottleneck Models (CBMs) to graphs is non-trivial. The unique characteristics of graphs pose challenges to the two key components of CBMs:
% (1) Predefining concepts: Unlike images, which often depict concrete objects (e.g. birds), graphs tend to represent abstract structures (e.g. social networks), making it more difficult to define and select meaningful, human-interpretable concepts.
% (2) Graph-to-concept alignment: It remains unclear how to effectively map the input domain (graphs) to the concept domain (languages). Moreover, due to the irregular structure and high variability of graphs across different domains, there is no readily available model—like CLIP~\cite{radford2021learningtransferablevisualmodels} in vision-language tasks—that can be directly applied for this alignment.

In light of these challenges, we propose \textbf{Graph Concept Bottleneck} (\model{}) as a new paradigm for interpretable text-attributed graph learning. \model{} consists of three modules. First, we pre-train a universal graph encoder using self-supervised \textit{contrastive concept–graph pretraining}, which aligns graph to the concept space. The encoder can be applied across different downstream datasets. Next, we initialize a concept set through \textit{LLM-enhanced concept retrieval} without manual annotation. This space is further refined by selecting the most informative concepts via \textit{information-constrained bottleneck optimization}. Finally, we train a predictor over the refined concept space to predict the label.
% \textcolor{blue}{In this paper, we focus on text-attributed graphs (TAGs). But we also discuss the feasibility for extending the method any other types of graphs.}

We conduct extensive experiments on five text-attributed graph datasets in clean and perturbed settings to evaluate \model{}. We highlight two major contributions of \model{}: (1) 
\textit{concept-based interpretable graph learning framework.} To the best of our knowledge, \model{} is the first self-explainable graph learning framework that projects graph inputs into a language space during prediction. This demonstrates the potential of actively incorporating natural language as an integral part of the reasoning process, enabling more interpretable and transparent deep learning on graph-structured data. (2)
\textit{Robust baseline for text-attributed node classification.} We show that \model{} performs competitively with SOTA GNNs in standard settings, and its advantages become more pronounced under distribution shifts and data perturbations, establishing a strong baseline for robust and generalizable graph learning.

% \begin{itemize}[leftmargin=1em, itemsep=0pt, topsep=0pt, parsep=0pt, partopsep=0pt]
%     \item \textbf{A language-based interpretable graph learning framework.} To the best of our knowledge, \model{} is the first self-explainable graph learning framework that projects graph inputs into a language space during prediction. This demonstrates the potential of actively incorporating natural language as an integral part of the reasoning process, enabling more interpretable and transparent deep learning on graph-structured data. Importantly, the explanations provided are faithful to the model’s decision process and accurately reflect the semantic meaning of the language concepts, without information leakage from labels to the concept space. 
    
%     \item \textbf{A robust baseline for node-level classification.} We show that \model{} performs competitively with state-of-the-art GNNs in standard settings, and its advantages become more pronounced under distribution shifts and data perturbations, establishing it as a strong baseline for robust and generalizable graph learning.
% \end{itemize}

 % To this end
 
\section{Related Work}
\label{sec::related_explainability}
In recent years, there has been growing interest in self-explainable GNNs~\cite{miao2022-GIB, miao2023-GIB, yu2020-GIB, yu2022-GIB, wu2020-GIB, wu2022-DIR, dai2021-SEGNN, feng2022-KERGNN, azzolin2025beyond, Wang2025prototype, liu2025faithfulclasslevelselfexplainabilitygraph, peng2024fewshotselfexplaininggraphneural}, where the explainability component is integrated into the prediction process. These methods typically generate informative subgraphs that serve both as explanations and as the basis for predictions. One line of work~\cite{miao2022-GIB, miao2023-GIB, yu2020-GIB, yu2022-GIB, wu2020-GIB} leverages the information bottleneck principle, aiming to extract the most informative yet compact subgraph by optimizing a graph information bottleneck objective. Other approaches~\cite{wu2022-DIR, dai2021-SEGNN, feng2022-KERGNN} introduce structural constraints to promote interpretability. For example, DIR~\cite{wu2022-DIR} decomposes the input graph into causal and non-causal components, enforcing that predictions depend only on the causal part. Despite these advances, most existing methods still rely on subgraphs as explanations, whose interpretability is not always guaranteed.
More recently, researchers have begun exploring alternative forms of explanation. For instance, \cite{bechler-speicher2024the} proposes Graph Neural Additive Networks, where the relationships between the input graph and the target variable can be directly visualized. \cite{sengupta2025xnodeselfexplanationneed} encodes interpretable cues (e.g. degrees, centrality) into a context vector, which is then mapped to an explanation vector. \cite{muller2023graphchef} employs decision trees to build rule-based predictors that are understandable to humans. However, none of these works employ natural language as a medium for explanations.

\section{Graph Concept Bottleneck}
\label{method}
Graph Concept Bottleneck consists of three stages: (1) we pretrain a graph-concept mapping model (\textbf{Section~\ref{sec::CCGP}}); (2) we create a concept set to explain the target dataset (\textbf{Section~\ref{sec::concept-space}} and \textbf{Section ~\ref{sec::ICBO}}); and (3) we train a predictor that maps concept activations to labels (\textbf{Section~\ref{sec::prediction}}). This multi-stage framework ensures that the model not only achieves strong predictive performance but also produces interpretable, concept-based explanations for its decisions.

% We introduce \textbf{Graph Concept Bottleneck (\model{})} as a new paradigm for interpretable graph learning. \model{} extracts semantic representations from graph data and maps them into a structured concept space, which serves as an intermediate layer providing human-understandable explanations for model predictions. 
% In this section, we detail the components of the \model{} framework. First, we pretrain graph encoders using a self-supervised contrastive learning objective that aligns graph representations with the concept space, enabling meaningful semantic correspondence between graphs and concepts (\textbf{Section ~\ref{sec::CCGP}}). Next, we develop the procedure to build the concept space. We initially construct a candidate concept set through a bi-level nomination procedure, combining both instance-based and global concept proposals (\textbf{Section ~\ref{sec::concept-space}}); To further refine the space, we select a subset of causal concepts using the \textit{Information Bottleneck criterion}, identifying concepts that are maximally informative about the target while being most compressed (\textbf{Section ~\ref{sec::ICBO}}). Finally, we train a predictor that leverages the graph-to-concept alignment scores of the selected concepts for downstream prediction tasks (\textbf{Section ~\ref{sec::prediction}}). 

% This multi-stage framework ensures that the model not only makes accurate predictions but also provides interpretable, concept-driven explanations of its decision process.

\subsection{Contrastive Concept–Graph Pretraining}

% \begin{wraptable}{r}{0.4\textwidth}
% \centering
% \scriptsize
% \vspace{-2em}
% \caption{F1 (\%) scores with GraphCLIP as the concept-graph alignment model, compared to GCN.
% }
% \vspace{-0.5em}
% \begin{tabular}{@{}lcc@{}}
% \toprule
% Model & \texttt{Cora} & \texttt{CiteSeer} \\
% \midrule

% GCN       & 72.38\textsubscript{(0.58)} & 63.81\textsubscript{(0.37)} \\
% GraphCLIP & 51.82\textsubscript{(0.65)} & 47.91\textsubscript{(1.76)} \\
% \bottomrule
% \end{tabular}
% \label{fig:graphclip}
% \vspace{-1em}
% \end{wraptable}

\label{sec::CCGP}
% Our goal is to have a pretrained multimodal model that can align graph and text representations to a shared space. A prior work on graph foundation models, GraphCLIP~\cite{zhu2025GraphCLIP}, pretrained a graph encoder and a text encoder by aligning graph inputs with text summaries generated by large language models (LLMs) using contrastive learning. However, directly integrating GraphCLIP into our framework results in substantially degraded performance relative to GCN, as shown in Figure~\ref{fig:graphclip}. We hypothesize that this is due to: (1) GraphCLIP aligns graphs to free-form summaries which contain noisy information, leading to inaccurate mappings between graphs and their underlying concepts; and (2) GraphCLIP jointly trains both the graph encoder and the text encoder, and the large number of parameters in the text encoder may lead to overfitting, especially when the training corpus is small or restricted to specific domains. To address these challenges, we propose a new scheme, Contrastive Concept–Graph Pretraining (CCGP). CCGP is specifically designed to enhance graph-to-concept alignment and can be universally applied to diverse datasets across different domains.
We propose Contrastive Concept–Graph Pretraining (CCGP), which pretrains a multimodal model to align graph and text representations in a shared space. CCGP is specifically designed to enhance graph-to-concept alignment and can be applied  across diverse datasets and domains.

% \begin{figure*}[t]
% \centerline{
% \includegraphics[scale=0.48]{images/c4_plot.pdf}
% }
% \caption{An overview figure of the proposed Graph Concept Bottleneck method in comparison with subgraph-based self-explainable graph neural networks.}
% \label{fig:intro}
% \end{figure*} 

\textbf{Pretraining data.} 
We collect unlabeled graph data from different domains to construct the pretraining datasets for CCGP. Prior work~\cite{wang2024llmszeroshotgraphlearners, chen2024llagalargelanguagegraph, tang2024graphgptgraphinstructiontuning} has demonstrated the remarkable ability of LLMs to understand and reason over graph-structured data. Motivated by this, we leverage LLMs to generate self-supervised concept annotations. For each dataset, we sample $m$ instances, where each instance $x_i$ corresponds to the ego-network centered at node $v_i$. For each instance $x_i$, we query GPT-3.5~\cite{brown2020gpt3} to generate a list of associated concepts (see Appendix~\ref{prompt:concept-annotation} for prompt details).
We collect instance–concept list pairs $\{(x_i, \mathcal{C}_i)\}$ for future procedures.
% \yue{what is graph here? an ID or name? what is dataset-details? list of nodes and edges?}

% \texttt{Given \{graphML\} and \{dataset-details\}. \setlist{nolistsep}
% \begin{enumerate} \itemsep0em 
% \item Provide summary and context analysis on the graph. \item Identify a list of key concepts and themes presented in the graph. \end{enumerate}
% }
% Here, \texttt{graphML} refers to the graph markup language used for describing the graph (or ego-net if the instance is a node). \texttt{dataset-details} provides a detailed description of the graph dataset, including what each node/edge represents and relevant contextual information. 

% \textbf{Data augmentation.}
We augment the pretrained data 
to improve the robustness of the pretrained model against data noise. For each instance $x_i$ we create a set of perturbed subgraphs  $\mathcal{X}_i^{\text{aug}} = \left\{ \tilde{x}_i^{(1)}, \tilde{x}_i^{(2)}, \dots, \tilde{x}_i^{(M)} \right\}$, 
% \begin{equation}
% \mathcal{X}_i^{\text{aug}} = \left\{ \tilde{x}_i^{(1)}, \tilde{x}_i^{(2)}, \dots, \tilde{x}_i^{(M)} \right\},
% \end{equation}
where each $\tilde{x}_i^{(m)}$ is constructed by perturbing the $k$-hop neighborhood of $v_i$ by randomly dropping/adding 20\% of the edges. The augmented instance–concept list pairs are obtained as $\{(\mathcal{X}_i^{\text{aug}}, \mathcal{C}_i)\}$

% One such augmented view is defined as:
% %
% \begin{equation}
% \tilde{x}_i^{(m)} = \mathcal{G} \left[
% \left( \mathcal{N}_k(v_i) \cup \mathcal{A}_i^{(m)}(\rho_{\text{add}}) \setminus \mathcal{R}_i^{(m)}(\rho_{\text{drop}}) \right),
% \left( \mathcal{E}_i \cup \mathcal{A}_i^{\text{edge}(m)}(\rho_{\text{add}}^{\text{edge}}) \setminus \mathcal{R}_i^{\text{edge}(m)}(\rho_{\text{drop}}^{\text{edge}}) \right)
% \right],
% \end{equation}

% where $\mathcal{A}_i^{(m)}$ and $\mathcal{R}_i^{(m)}$ are the sets of added and removed nodes for the $m$-th augmentation, sampled according to the perturbation parameters $\rho_{\text{add}}$ and $\rho_{\text{drop}}$. Similarly, $\mathcal{A}_i^{\text{edge}(m)}$ and $\mathcal{R}_i^{\text{edge}(m)}$ represent edge perturbations. 
% We obtain the augmented instance–concept list pairs, denoted as $\{(\mathcal{X}_i^{\text{aug}}, \mathcal{C}_i)\}$.

\textbf{Encoders.} The pretrained model consists of a graph encoder and a text encoder. 
The graph encoder $f_{\theta}^{\text{GNN}}(\cdot)$ with trainable parameter $\theta$ is responsible for capturing both the feature attributes and topological structure of the graph, and it should generalize well to downstream graph data, potentially from different datasets.
More expressive architectures, such as Graph Transformers, are capable of modeling rich semantics and complex patterns, but they are more prone to overfitting than smaller GNNs like GCN.
%A practical rule of thumb is to select a graph encoder that demonstrates consistently strong performance across most downstream datasets within the target domain.
% We denote the graph encoder as $f_{\theta}^{\text{GNN}}(\cdot)$, where $\theta$ represents the learnable parameters of the graph encoder. 
We adopt a pretrained Sentence-BERT~\cite{reimers-2019-sentence-bert} model as the text encoder $f^{\text{LM}}(\cdot)$, with parameters kept frozen throughout training. This choice leverages the model's strong general semantic capabilities, while avoiding the computational cost and overfitting risks associated with fine-tuning large language models on limited data.

\textbf{Set2set contrastive learning.}
Unlike standard contrastive learning, which aligns the input with a single target, our setting involves multiple augmented graph views and  concepts per instance. We thus formulate graph–concept pretraining as a \emph{set-to-set} alignment problem, encouraging the model to consistently associate diverse structural perturbations of a graph with all of its semantic concepts.

For each instance $x_i$, we have a set of augmented views $\mathcal{X}_i^{\text{aug}}=\{\tilde{x}_i^{(1)},\dots,\tilde{x}_i^{(M)}\}$ and a concept set $\mathcal{C}_i=\{c_{i1},\dots,c_{i2},\dots,c_{iK}\}$. During training, we sample positive pairs from their Cartesian product,
\[
P_i=\{(\tilde{x}_i^{(m)},c_{ij})\mid m\in\mathcal{M}_i,\; j\in\mathcal{K}_i\},
\]
where $\mathcal{M}_i\subseteq\{1,\dots,M\}$ and $\mathcal{K}_i\subseteq\{1,\dots,K\}$ are sampled subsets. For each $(\tilde{x}_i^{(m)},c_{ij})\in P_i$, we compute the graph embedding $z_i^{(m)}=f_{\theta}^{\text{GNN}}(\tilde{x}_i^{(m)})$ and the concept embedding $z_{ij}^{\text{concept}}=f^{\text{LM}}(c_{ij})$.
We then apply a contrastive loss based on the InfoNCE~\cite{oord2018representation} formulation to maximize the similarity between positive pairs while minimizing similarity to negative pairs in the batch. The contrastive loss for each positive pair is defined as:
\vspace{-1.5mm}
\begin{equation}
\mathcal{L}_{i,j}^{(m)} = -\log \frac{
\exp\left( \text{sim}\left( z_i^{(m)}, z_{ij}^{\text{concept}} \right) / \tau \right)
}{
\sum\limits_{\substack{(k, l) \in \mathcal{B}}} 
\exp\left( \text{sim}\left( z_i^{(m)}, z_{kl}^{\text{concept}} \right) / \tau \right)
},
\end{equation}
where $\text{sim}(\cdot, \cdot)$ is cosine similarity, $\tau$ is the temperature, and $\mathcal{B}$ is the set of all (view, concept) pairs in the current batch.
Overall, we formulate the learning objective as minimizing the following contrastive loss with respect to \(\theta\):
\vspace{-2mm}
{\setlength{\abovedisplayskip}{8pt}
 \setlength{\belowdisplayskip}{8pt}
 \begin{equation}
 \arg\min_{\theta}\,\mathcal{L}(\theta)
 = \frac{1}{\sum_i |\mathcal{M}_i|\,|\mathcal{K}_i|}
 \sum_i\sum_{m\in\mathcal{M}_i}\sum_{j\in\mathcal{K}_i}
 \mathcal{L}_{i,j}^{(m)}(\theta).
 \end{equation}
}

By doing so, the model learns to align multiple augmented views of each graph with multiple concepts, improving generalization across diverse graph data. We denote the optimized graph encoder as $f^{\text{GNN}}(\cdot)$, whose parameters are frozen during subsequent training to ensure independence from label supervision, thereby preserving semantic meaning and avoiding information leakage.

\subsection{LLM-Empowered Concept Retrieval}
\label{sec::concept-space}

% Inspired by Oikarinen et al~\cite{oikarinen2023labelfreeconceptbottleneckmodels}., we leverage LLMs~\cite{radford2019gpt2, brown2020gpt3, grattafiori2024llama3herdmodels, deepseekai2025deepseekv3technicalreport} to harvest candidate concepts.
Given the strong ability of LLMs in domain knowledge~\cite{lee2024LLMdomain}, abstraction $\&$ pattern recognition~\cite{lee2024LLMreasoning}, and contextual reasoning~\cite{zhang2024llmreasoning}, we construct the concept space through two approaches:

\textbf{Global Concept Proposal.}
We expect LLMs to identify concepts to distinguish between classes when instructed appropriately. Specifically, we provide detailed description of the dataset and ask the LLM to generated an initial set of revelent concepts for each class. See ~\ref{prompt:global-concept} for prompt details.
% We ask GPT-3.5~\cite{brown2020gpt3} the following:
% \texttt{In the domain of \{dataset-domain\}, list the related concepts/keywords for classifying the item as \{category\}.} 
% Here, \texttt{dataset-domain} briefly describes the dataset’s domain or context, and \texttt{category} is the name of a class label from the downstream classification task. 
% We apply this prompt to each class label and aggregate the generated concepts to form the initial concept pool.

\textbf{Instance-Based Concept Extraction. }
While Global Concept Proposal offers broader semantic coverage and reflects domain-level priors, it may overlook dataset-specific nuances that can only accessible through instances.
Observing the richness of the training data and LLMs' ability to perform contextual reasoning and summarize fine-grained patterns on graph data~\cite{wang2024llmszeroshotgraphlearners, chen2024llagalargelanguagegraph, tang2024graphgptgraphinstructiontuning}, we propose to ask LLMs to recognize relevant concepts given sampled graph instances.
% (see Appendix~\ref{prompt::local-concept} for prompt details). 
% Specifically, we instruct GPT-3.5~\cite{brown2020gpt3} the following:
% \texttt{Given a \{\texttt{graphML}\} and \{dataset-details\}. \setlist{nolistsep}
% \begin{enumerate}[noitemsep]
% \item Provide summary and context analysis on the graph. \item Identify a list of key concepts presented in the graph that are most important for determining its classification within the \{dataset-domain\}, which includes the following categories: \{category-list\}. \end{enumerate}
% }
% The query format here follows a similar structure to the data preparation process described in Section~\ref{sec::CCGP}, with one key difference: we include the complete list of categories for the classification task, denoted as \texttt{category-list}, to guide the LLM toward generating concepts that are helpful in predicting class labels. Note that only the outputted concept list from the second step is collected. However, we find that explicitly prompting the LLM to first analyze the graph and produce a summary helps it better understand the context, thereby improving the quality of the generated concepts.
Specifically, we sample $m$ graph instances from each class and apply the prompt to each sampled graph instance, resulting in a large set of candidate concepts. We then identify a subset of concepts that are highly relevant to each class, distinct from those used by other classes, and useful for improving class discrimination. 
% See Appendix \ref{append::instance-based} for the details of this process. 
% Specifically, for each class $y$, we calculate the class-wise concept activation score as:
% $\bar{C}_y = \frac{1}{|\mathcal{D}_y|} \sum_{x_i \in \mathcal{D}_y} C_i$,
% where $\mathcal{D}_y$ denotes the set of instances belonging to class $y$, and $C_i$ is the concept activation vector for instance $x_i$. Each element $C_i^{(j)}$ represents the activation score (e.g., cosine similarity) between the instance representation $f_{\theta}^{\text{GNN}}(x_i)$ and the embedding of the $j$-th concept $f^{\text{LM}}(c_j)$.
% We then compute the discriminative score of concept $j$ for class $y$ as:
% $ \text{score}_j(y) = \bar{C}_y^{(j)} - \frac{1}{|\mathcal{Y}| - 1} \sum_{y' \ne y} \bar{C}_{y'}^{(j)},
% $
% where $\mathcal{Y}$ is the set of all class labels, and $\bar{C}_y^{(j)}$ denotes the average activation of concept $j$ for class $y$. 
% Finally, for each class, we select the top-$k$ concepts with the highest discriminative scores:
% \begin{equation}
%     \mathcal{C}^{\text{inst}} = \text{Top-$k$}_{j}(\text{score}_j(y)).
% \end{equation}
%
To control the quality and size of the concept set we perform several filtering steps to remove redundant or irrelevant concepts.
Details of the prompting and filtering process are provided in ~\ref{append::filtering}. We denote the filtered concepts from the Global Concept Proposal and Instance-Based Concept Extraction $\mathcal{C}^{\text{glob}}$ and $\mathcal{C}^{\text{inst}}$, separately. 
We combine $\mathcal{C}^{\text{glob}}$ and $\mathcal{C}^{\text{inst}}$ as the candidate concept set as $\mathcal{C}^{\text{candidate}}$. 
% We set $k$ to be 100. 

\subsection{Information-Constrained Concept Optimization}
\label{sec::ICBO}

The retrieved concept set in Section \ref{sec::concept-space} is potentially large, and some of them could be irrelevant or spurious, hindering the explanablity and the generalizability of the model. Thus, we adopt the Information Bottleneck (IB)~\cite{alemi2019IB} criteria to encourage the model to rely on a sparse set of concepts that are most relevant to the prediction.  

% \textbf{Information Bottleneck}~\cite{alemi2019IB}. 
\begin{definition}
    The Information Bottleneck criteria is generally formulated as $I(Z; Y) - \beta I(Z; X)$, which seeks a representation $Z$ that is both \textit{informative} and \textit{compressed}: maximizing mutual information with the label $Y$ while minimizing mutual information with the input $X$. A larger $\beta$ results in stronger compression, encouraging $Z$ to retain only the most essential information for predicting $Y$.
\end{definition}

In our model, we adopt the IB objective to learn a gating vector \( g \) over the fixed concept space. Specifically, for each concept \( j \), we learn a soft gate \( \mathbf{g}_j \) by applying a sigmoid function to a learnable parameter.

% \vspace{-1.5mm}
% \begin{equation}
% \textbf{g}_j = \sigma\big(\text{MLP}^{\text{gate}}_{\phi}(f^{\text{LM}}(c_j))\big),
% \end{equation}
% where $\text{MLP}^{\text{gate}}_{\phi}(\cdot)$ is a learnable multi-layer perceptron applied to the concept embedding $f^{\text{LM}}(c_j)$, 
% and $\sigma(\cdot)$ denotes the sigmoid activation function. 
We then apply the gate vector to the concept activation vector of each instance $i$ as $\textbf{z}_{i} = \textbf{g} \odot \textbf{a}_{i}$, where $\odot$ is element-wise multiplication, $\textbf{a}_i$ is the concept activation vector for instance $i$, defined as:
\vspace{-2.5mm}
\begin{equation}
\label{eq:activation-vector}
\mathbf{a}_i^{(j)} = \text{sim} \left( f^{\text{GNN}}(x_i),\ f^{\text{LM}}(c_j) \right),
\end{equation}
where \( \text{sim}(\cdot, \cdot) \) is the cosine similarity.
$\textbf{z}_i$ is the masked concept vector passed to the classifier.
Following the IB principle, we minimize the following objective:
\vspace{-1.5mm}
\begin{equation}
\frac{1}{N} \sum_{i=1}^{N} 
\mathbb{E} \left[
    -\log q(y_i |\textbf{z}_{i})
\right]
+ \beta \, \text{KL}\left( p(\textbf{z}_{i}|\textbf{a}_i) \, \| \, r(\textbf{z}_{i}) \right),
\label{eq:ib-objective}
\end{equation}
where the first term promotes predictive accuracy and the second term minimizes the Kullback–Leibler (KL) divergence between $\textbf{z}_i$ and  $\textbf{x}_i$, effectively penalizing their mutual information and encouraging a more compressed representation.
In our framework, the prediction function \( q(y_i|\textbf{z}_i) \) is parameterized by a trainable multi-layer perceptron \( \text{MLP}^{\text{cls}}_{\psi} \), which takes the masked concept vector \( z_i \) as input. The gate vector \( \textbf{g} \), which determines the masking over the concept activations, is computed by a separate network parameterized by \( \phi \).
Since \( \textbf{z}_i \) is deterministically computed and we do not model a distribution over \( p(\textbf{z}_i|x_i) \), we approximate the KL divergence term with a deterministic sparsity regularizer. In particular, we use an \( L_1 \) penalty, which encourages the gate values to shrink toward zero, effectively suppressing irrelevant concepts. This results in a sparse, interpretable concept selection, aligning with the Information Bottleneck's objective of compressing the intermediate representation while retaining task-relevant information.
Thus, the training objective in Equation \ref{eq:ib-objective} becomes:
\vspace{-2.5mm}
\begin{equation}
\min_{\phi, \psi} \;
\frac{1}{N} \sum_{i=1}^{N} 
\mathcal{L}_{\text{CE}}\left( \text{MLP}^{\text{cls}}_{\psi}(\textbf{z}_i), \, y_i \right)
+ \beta \, \|\textbf{g}\|_1,
\label{eq:ib-objective-in-practice}
\end{equation}
where \( \mathcal{L}_{\text{CE}} \) denotes the cross-entropy loss between the predicted label distribution and the ground-truth label.
%The Information Bottleneck objective is used solely to train the gate vector $g$. 
While the gates are continuous and soft during training, for interpretability, we require a discrete selection of concepts. To achieve this, after the IB training phase, we freeze the learned gate vector $\mathbf{g}$ and select the top-$K$ concepts with the highest gate values 
$
\mathcal{C}^{\text{selected}} = \text{Top-K}_{j}(\textbf{g}_j),
$
where $\mathcal{C}^{\text{selected}}$ denotes the final set of selected concepts.

\subsection{Predictor Learning}
\label{sec::prediction}

We use the selected concept set \( \mathcal{C}^{\text{selected}} \) to make the final predictions. For each instance \( x_i \), we compute a concept activation vector \( \mathbf{a}_i^{\mathcal{C}} \in \mathbb{R}^{|\mathcal{C}^{\text{selected}}|} \) following Equation~\ref{eq:activation-vector}, where each dimension corresponds to a concept in \( \mathcal{C}^{\text{selected}} \).
We then train a predictor using only the concept activation vector \( \mathbf{a}_i^{\mathcal{C}} \) as input. Specifically, we use a classifier \( \text{MLP}^{\text{cls}}_{\theta} \), parameterized by \( \theta \) to predict the label \( y_i \). The objective is to minimize the loss over the training set:
\vspace{-2.5mm}
\begin{equation}
\theta^\ast = \arg\min_{\theta} \ \frac{1}{N} \sum_{i=1}^{N} \mathcal{L}_{\text{CE}}\left( \text{MLP}^{\text{cls}}_{\theta}\left( \mathbf{a}_i^{\mathcal{C}} \right),\ y_i \right),
\end{equation}
where \( \mathcal{L}_{\text{CE}} \) denotes the standard cross-entropy loss, and \( N \) is the number of training instances.

% \subsection{Discussion}
% \textbf{Information Leakage. } The leakage of label information into the concepts is a major concern in concept learning models, as it can severely undermine their interpretability and faithfulness~\cite{Havasi2022, sun2024eliminatinginfo}. When the CLM is trained jointly, the label predictor may exploit spurious signals from the concept activations produced by the concept predictor, rather than relying on the true semantics of the concepts. In other words, even if the concepts are meaningless or unrelated, the model may still achieve high accuracy by learning to assign higher activation scores to random concepts whenever it recognizes the label of an instance, while these concepts provide no real explainable value. 

% We point out that \model{} suffers minimal potential leakage due to the task-blind~\cite{mahinpei2021spitfall} nature of its concept predictor. The concept predictor and the label predictor are trained independently rather than jointly: the concept predictor is first pretrained, and its output concept activations are then used as fixed inputs when training the label predictor. To further prevent label information leakage during concept predictor pretraining with CCGP, we exclude concepts that are semantically similar to the label itself during the concept retrieval process. In addition, we evaluate \model{} on datasets from disjoint domains when (pre)training the concept predictor and the label predictor. These design choices collectively minimize the risk of information leakage.
\section{Experiments}
\label{experiments}

Given the self-explainable nature of \model{}, we evaluate it from two aspects: as a \emph{predictor} and as an \emph{explainer}. Specifically, we investigate whether \model{} can make robust predictions on text-attributed node classification under distribution shift and data perturbations, meanwhile, whether the explanations induced by its concepts are faithful to the model’s decision process, sufficient and necessary for prediction, and concise enough to be easily interpretable by humans.
% We investigate the \textit{robustness} and \textit{interpretability} of \model{}. First, we evaluate its utility across datasets from different domains and under varying conditions to assess robustness and generalizability to distribution shifts and data perturbations (see Section~\ref{sec:robustness}). We then closely examine the faithness of the concepts and investigate whether there exist information leakage (Section~\ref{sec::info_leakge}). We then analyze the sensitivity of \model{} to concept size and the choice of graph encoders (Section~\ref{sec::sensitivity}).  Finally, we conduct a case study to visualize how \model{} provides intuitive explanations for its predictions via the concept bottleneck layer (Section~\ref{sec::interpretability}).

% answer the following research questions (\textbf{RQs}). \textbf{RQ1 - Utility:} How accurately can \model{} make predictions, and how robust is it under distribution shifts and data perturbations? 
% \textbf{RQ2 - Interpretability:} How does the concept bottleneck provide explanations to the 

% \textbf{RQ3 - Sensitivity:} How sensitive is \model{} to different hyperparameters?
\begin{table*}[h]
    \centering
    \fontsize{7.5pt}{7.5pt}\selectfont
    \caption{Node classification performance in \textit{OOD settings} with upsampling ratio $\gamma=5$. The best-performing interpretable GNN is \underline{underlined}, and the overall best-performing method is \textbf{bolded}.}% (averaged over 5 trials). Standard deviation is denoted after $\pm$.}
\label{tab:res_ood}
\begin{tabular}{l@{\hskip 5pt}c@{\hskip 3pt}c@{\hskip 7pt}@{\hskip 0pt}c@{\hskip 3pt}c@{\hskip 7pt}@{\hskip 0pt}c@{\hskip 3pt}c@{\hskip 7pt}@{\hskip 0pt}c@{\hskip 3pt}c@{\hskip 7pt}@{\hskip 0pt}c@{\hskip 3pt}c}
  % \begin{tabular}{lcccccccccc}
\toprule
 & \multicolumn{2}{c}{\texttt{Cora}} & \multicolumn{2}{c}{\texttt{Citeseer}} & \multicolumn{2}{c}{\texttt{Instagram}} & \multicolumn{2}{c}{\texttt{Reddit}} & \multicolumn{2}{c}{\texttt{WikiCS}} \\
 \cmidrule(lr){2-3} \cmidrule(lr){4-5} \cmidrule(lr){6-7} \cmidrule(lr){8-9} \cmidrule(lr){10-11} 
\textbf{Method} & F1 (\%) & BACC (\%) & F1 (\%) & BACC (\%) & F1 (\%) & BACC (\%) & F1 (\%) & BACC (\%) & F1 (\%) & BACC (\%) \\
\midrule
MLP & 50.00\textsubscript{(0.63)} & 60.81\textsubscript{(0.50)} & 38.44\textsubscript{(0.99)} & 54.34\textsubscript{(1.10)} & 35.42\textsubscript{(0.58)} & 51.69\textsubscript{(0.26)} & 16.38\textsubscript{(0.52)} & 51.33\textsubscript{(0.51)} & 54.31\textsubscript{(0.31)} & 65.30\textsubscript{(0.35)} \\
GCN & 55.10\textsubscript{(0.66)} & 64.03\textsubscript{(1.05)} & 46.52\textsubscript{(0.92)} & 59.64\textsubscript{(0.92)} & 36.98\textsubscript{(0.84)} & 50.01\textsubscript{(0.70)} & 12.91\textsubscript{(0.38)} & 48.48\textsubscript{(0.21)} & \textbf{59.04}\textsubscript{(1.66)} & \textbf{68.38}\textsubscript{(1.73)} \\
GAT & 51.30\textsubscript{(1.27)} & 61.52\textsubscript{(1.15)} & 45.62\textsubscript{(1.01)} & 58.77\textsubscript{(0.87)} & 33.31\textsubscript{(0.59)} & 50.18\textsubscript{(0.41)} & 12.93\textsubscript{(0.30)} & 49.34\textsubscript{(0.18)} & 57.05\textsubscript{(1.00)} & 64.53\textsubscript{(0.85)} \\
SAGE & 44.26\textsubscript{(1.78)} & 53.95\textsubscript{(1.65)} & 30.87\textsubscript{(0.41)} & 48.42\textsubscript{(0.47)} & 31.46\textsubscript{(0.20)} & 48.27\textsubscript{(0.18)} & 13.38\textsubscript{(0.17)} & 49.18\textsubscript{(0.47)} & 51.87\textsubscript{(1.24)} & 62.06\textsubscript{(1.37)} \\
GT & 38.26\textsubscript{(1.64)} & 48.66\textsubscript{(1.36)} & 28.38\textsubscript{(1.58)} & 48.22\textsubscript{(0.77)} & 30.90\textsubscript{(0.43)} & 48.06\textsubscript{(0.37)} & 12.64\textsubscript{(0.54)} & 48.62\textsubscript{(0.26)} & 54.06\textsubscript{(0.84)} & 62.50\textsubscript{(0.77)} \\
\midrule
DIR-GNN & 23.07\textsubscript{(2.70)} & 43.18\textsubscript{(2.13)} & 15.31\textsubscript{(1.33)} & 42.93\textsubscript{(1.00)} & 26.74\textsubscript{(0.00)} & 50.00\textsubscript{(0.00)} & 8.46\textsubscript{(0.00)} & 50.00\textsubscript{(0.00)} & 22.93\textsubscript{(1.35)} & 42.11\textsubscript{(0.42)} \\
GIB & 19.23\textsubscript{(4.10)} & 40.24\textsubscript{(3.48)} & 15.52\textsubscript{(1.79)} & 42.31\textsubscript{(1.19)} & 26.75\textsubscript{(0.01)} & 50.00\textsubscript{(0.01)} & 8.47\textsubscript{(0.03)} & 50.01\textsubscript{(0.01)} & 24.98\textsubscript{(1.27)} & 39.15\textsubscript{(1.12)} \\
VGIB & 44.56\textsubscript{(6.43)} & 57.06\textsubscript{(4.66)} & 22.26\textsubscript{(6.35)} & 47.72\textsubscript{(3.26)} & 26.74\textsubscript{(0.00)} & 50.00\textsubscript{(0.00)} & 8.46\textsubscript{(0.00)} & 50.00\textsubscript{(0.00)} & 56.02\textsubscript{(1.76)} & 64.14\textsubscript{(1.25)} \\
SEGNN & 30.68\textsubscript{(2.91)} & 48.75\textsubscript{(1.96)} & 19.92\textsubscript{(2.80)} & 42.89\textsubscript{(1.55)} & 26.74\textsubscript{(0.00)} & 50.00\textsubscript{(0.00)} & 8.46\textsubscript{(0.00)} & 50.00\textsubscript{(0.00)} & 34.97\textsubscript{(1.26)} & 50.71\textsubscript{(1.01)} \\
\model{} & \underline{\textbf{56.63}}\textsubscript{(1.38)} & \underline{\textbf{66.71}}\textsubscript{(0.99)} & \underline{\textbf{60.19}}\textsubscript{(0.61)} & \underline{\textbf{67.12}}\textsubscript{(0.60)} & \underline{\textbf{56.80}}\textsubscript{(0.23)} & \underline{\textbf{58.47}}\textsubscript{(0.38)} & \underline{\textbf{48.16}}\textsubscript{(0.25)} & \underline{\textbf{63.07}}\textsubscript{(0.99)} & \underline{56.36}\textsubscript{(0.46)} & \underline{67.57}\textsubscript{(0.73)} \\
\bottomrule
\end{tabular}
\end{table*}

\begin{table*}[h]
    \centering
    % \scriptsize
    \fontsize{7.5pt}{7.5pt}\selectfont
        \caption{Node classification performance in \textit{perturbation settings} with pertubation ratio $\rho=0.3$. The best-performing interpretable GNN is \underline{underlined}, and the overall best-performing method is \textbf{bolded}.}% (averaged over 5 trials). Standard deviation is denoted after $\pm$.}
\label{tab:res_pertube}
\begin{tabular}{l@{\hskip 5pt}c@{\hskip 3pt}c@{\hskip 7pt}@{\hskip 0pt}c@{\hskip 3pt}c@{\hskip 7pt}@{\hskip 0pt}c@{\hskip 3pt}c@{\hskip 7pt}@{\hskip 0pt}c@{\hskip 3pt}c@{\hskip 7pt}@{\hskip 0pt}c@{\hskip 3pt}c}
\toprule
 & \multicolumn{2}{c}{\texttt{Cora}} & \multicolumn{2}{c}{\texttt{Citeseer}} & \multicolumn{2}{c}{\texttt{Instagram}} & \multicolumn{2}{c}{\texttt{Reddit}} & \multicolumn{2}{c}{\texttt{WikiCS}} \\
 \cmidrule(lr){2-3} \cmidrule(lr){4-5} \cmidrule(lr){6-7} \cmidrule(lr){8-9} \cmidrule(lr){10-11} 
\textbf{Method} & F1 (\%) & BACC (\%) & F1 (\%) & BACC (\%) & F1 (\%) & BACC (\%) & F1 (\%) & BACC (\%) & F1 (\%) & BACC (\%) \\
\midrule
MLP & 43.08\textsubscript{(0.45)} & 56.55\textsubscript{(0.37)} & 37.77\textsubscript{(0.53)} & 53.77\textsubscript{(0.59)} & 36.69\textsubscript{(0.45)} & 52.57\textsubscript{(0.19)} & 16.75\textsubscript{(0.84)} & 51.11\textsubscript{(0.44)} & 53.70\textsubscript{(0.32)} & 64.95\textsubscript{(0.45)} \\
GCN & 56.21\textsubscript{(1.25)} & 66.49\textsubscript{(0.85)} & 46.45\textsubscript{(1.42)} & 58.14\textsubscript{(1.61)} & 43.87\textsubscript{(3.28)} & 54.11\textsubscript{(0.85)} & 16.64\textsubscript{(0.77)} & 50.43\textsubscript{(0.76)} & 61.45\textsubscript{(0.38)} & 65.64\textsubscript{(0.87)} \\
GAT & 52.64\textsubscript{(1.33)} & 61.70\textsubscript{(0.71)} & 46.55\textsubscript{(0.87)} & 60.54\textsubscript{(0.58)} & 37.07\textsubscript{(1.09)} & 52.38\textsubscript{(0.43)} & 19.23\textsubscript{(1.76)} & 52.20\textsubscript{(0.84)} & 59.34\textsubscript{(0.90)} & 66.75\textsubscript{(1.21)} \\
SAGE & 53.15\textsubscript{(2.06)} & 57.70\textsubscript{(2.21)} & 39.37\textsubscript{(1.30)} & 56.60\textsubscript{(0.90)} & 38.45\textsubscript{(2.14)} & 52.79\textsubscript{(0.90)} & 16.51\textsubscript{(0.49)} & 51.64\textsubscript{(0.39)} & 61.58\textsubscript{(0.44)} & 69.76\textsubscript{(0.49)} \\
GT & 46.13\textsubscript{(2.23)} & 55.15\textsubscript{(1.80)} & 33.36\textsubscript{(1.69)} & 52.65\textsubscript{(0.93)} & 35.36\textsubscript{(0.78)} & 51.70\textsubscript{(0.27)} & 17.29\textsubscript{(0.50)} & 51.31\textsubscript{(0.12)} & 57.88\textsubscript{(0.95)} & 62.61\textsubscript{(1.60)} \\
\midrule
DIR-GNN & 70.78\textsubscript{(2.43)} & 70.20\textsubscript{(2.49)} & 62.03\textsubscript{(0.86)} & \underline{\textbf{64.65}}\textsubscript{(0.70)} & 55.56\textsubscript{(1.43)} & 56.45\textsubscript{(0.64)} & 54.13\textsubscript{(1.52)} & \underline{\textbf{56.35}}\textsubscript{(0.57)} & 57.07\textsubscript{(3.45)} & 56.34\textsubscript{(1.97)} \\
GIB & 32.94\textsubscript{(18.33)} & 37.41\textsubscript{(15.43)} & 47.23\textsubscript{(15.64)} & 52.91\textsubscript{(11.58)} & 38.55\textsubscript{(6.36)} & 50.74\textsubscript{(0.95)} & 39.96\textsubscript{(7.82)} & 51.54\textsubscript{(1.85)} & 21.62\textsubscript{(10.37)} & 25.78\textsubscript{(9.43)} \\
VGIB & 20.15\textsubscript{(26.58)} & 26.24\textsubscript{(23.92)} & 54.92\textsubscript{(20.60)} & 57.43\textsubscript{(17.87)} & 39.13\textsubscript{(0.61)} & 50.09\textsubscript{(0.17)} & 34.79\textsubscript{(3.10)} & 50.20\textsubscript{(0.39)} & 58.67\textsubscript{(24.39)} & 59.70\textsubscript{(22.48)} \\
SEGNN & 52.58\textsubscript{(4.71)} & 56.78\textsubscript{(3.34)} & 59.76\textsubscript{(1.11)} & 62.69\textsubscript{(1.06)} & 55.15\textsubscript{(0.64)} & 55.40\textsubscript{(0.43)} & 55.44\textsubscript{(0.85)} & 55.77\textsubscript{(0.71)} & 38.08\textsubscript{(1.10)} & 42.11\textsubscript{(1.15)} \\
\model{} & \underline{\textbf{70.98}}\textsubscript{(0.73)} & \underline{\textbf{71.36}}\textsubscript{(1.07)} & \underline{\textbf{63.44}}\textsubscript{(0.29)} & 63.84\textsubscript{(0.32)} & \underline{\textbf{56.65}}\textsubscript{(0.26)} & \underline{\textbf{56.61}}\textsubscript{(0.27)} & \underline{\textbf{55.56}}\textsubscript{(0.74)} & 55.58\textsubscript{(0.77)} & \underline{\textbf{66.08}}\textsubscript{(0.53)} & \underline{\textbf{70.43}}\textsubscript{(0.75)} \\
\bottomrule
\end{tabular}
\end{table*}
\subsection{Datasets} \label{Sec:data}

Following the practice in GraphCLIP~\citep{zhu2025GraphCLIP}, we use non-overlapping datasets from diverse domains to pretrain the Graph-Concept Alignment model. The graph data used for pre-training is required to be of the same type as the downstream datasets to ensure transferability. In this work, we focus on \textit{text-attributed graphs}, where node attributes are textual descriptions of their contents.

\textbf{Source datasets.}
We use five source datasets for CCGP: \texttt{Pubmed}~\citep{Sen2008cora} is citation network in medical domain. \texttt{Ele-Computers}, \texttt{Sports-Fitness}, \texttt{Books-Children}, and \texttt{Books-History}~\citep{Yan2023TAG} are e-commerce co-purchasing networks. 
For each dataset, we sample 1{,}000 nodes and query GPT-3.5 Turbo to generate 10 concepts that appear in each node's ego network, serving as labels for the graph-concept alignment.

\textbf{Target datasets.}
We use \texttt{Cora}~\citep{Sen2008cora}, \texttt{Citeseer}~\citep{Sen2008cora}, \texttt{Instagram}~\citep{huang2024Can}, \texttt{Reddit}~\citep{huang2024Can}, and \texttt{WikiCS}~\citep{mernyei2020wiki} as target datasets. \texttt{Cora} and \texttt{Citeseer} are citation networks in Computer Science; \texttt{Instagram} and \texttt{Reddit} are social networks; and \texttt{WikiCS} is a Wikipedia article network. \textit{We ensure that all target datasets are from different domains than the source datasets to evaluate model generalizability and reduce data leakage.}
We evaluate them under three settings:

\begin{itemize}[leftmargin=1em, itemsep=0pt, topsep=0pt, parsep=0pt, partopsep=0pt]
    \item \textit{Regular.} We randomly split data into training, validation, and test sets that follow the same data distribution. 
    % \yue{can you explain inductive setting here?}
\item \textit{OOD.} Following the \textit{soft label-leaveout} strategy~\citep{han2024decaf}, we create mismatched class distributions across splits using an upsampling ratio $\gamma$. Specifically, classes are partitioned into majority and minority groups. Instances from majority classes are then $\gamma$ times more likely to be sampled into the training/validation set than those from minority classes, while keeping the overall training/validation/test split ratios unchanged. A larger $\gamma$ therefore induces a stronger distribution shift between the training and test sets.
 We set $\gamma \in \{2, 3, 5, 10\}$.
    
\item \textit{Adversarial.} Using the same split as the \textit{regular setting}, we perturb the edges in the training set by randomly dropping and adding edges to the graph with a perturbation ratio $\rho$. We set $\rho \in \{0.05, 0.1, 0.2, 0.3, 0.5\}$.
\end{itemize}

% For all datasets and settings, we adopt a default train/validation/test split of 20\%/20\%/50\%, with the remaining 10\% held out to investigate the effect of increased training data in later experiments. We use an \textit{inductive} splitting, where test nodes are entirely unseen during training and vice versa.
We report the Macro F1 scores and Balanced Accuracy Score (BACC) to evaluate model performance to account for class imbalance. See \ref{sec::datasets} and \ref{sec::implementations} for further details on the datasets and experimental settings.

\subsection{\model{} as a Predictor}
\label{sec:robustness}
We evaluate \model{} as a predictive model on target datasets under three different settings. We set the default number of concepts of \model{} $K$ as 30. For each setting, we also evaluate a set of SOTA GNN and MLP models, including MLP, GCN~\cite{kipf2017GCN}, GAT~\cite{veličković2018GAT}, GraphSAGE (SAGE)~\cite{hamilton2018SAGE}, and Graph Transformer (GT)~\cite{yun2020GT}, as baselines. We also compare with self-explainable GNNs including GIB~\cite{yu2020-GIB}, VGIB~\cite{yu2022-GIB}, DIRGNN~\cite{wu2022-DIR}, and SEGNN~\cite{dai2021-SEGNN}. 
% \yue{Introduce the baselines for self-explainable GNNs (GIB, VGIB, DIRGNN)}

 \begin{figure*}[t]
  \centering
    \includegraphics[width=\textwidth]{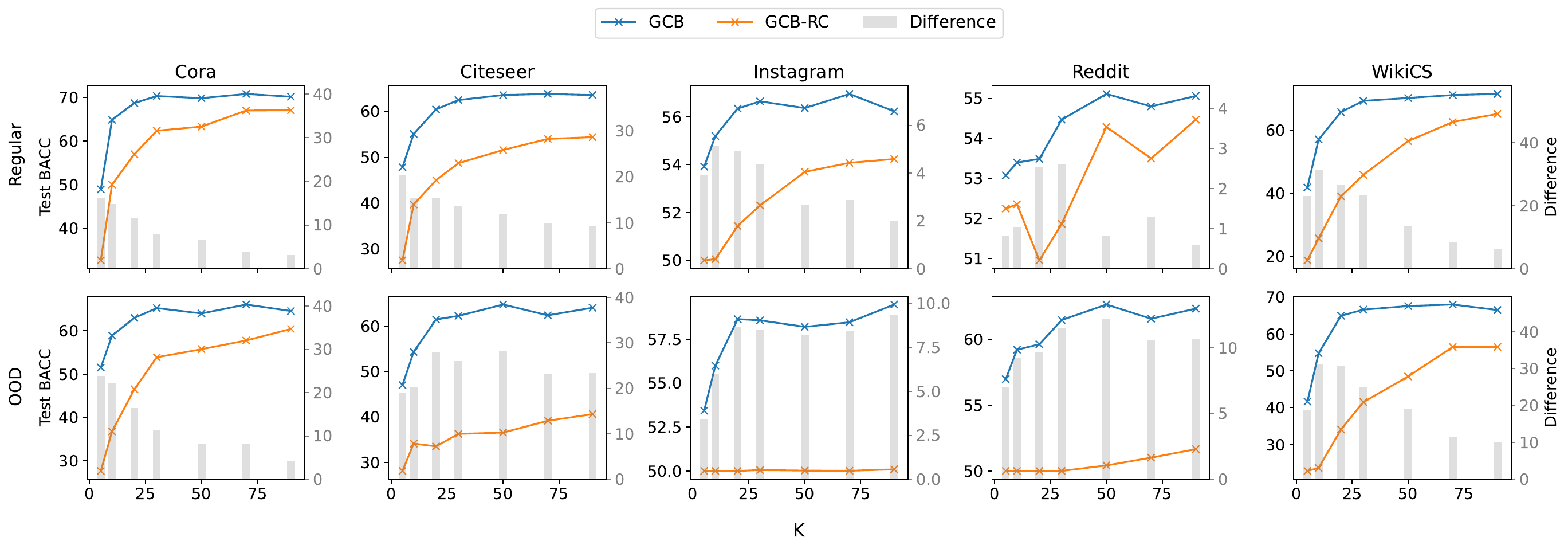}
  \caption{Performance of the original \model{} compared to its variant with random concepts (GCB-RC) across different concept set sizes, on regular splits (top row) and OOD splits (bottom row).}
  \label{fig:rc}
\end{figure*}

\begin{figure*}[h]
  \centering
    \includegraphics[width=\textwidth]{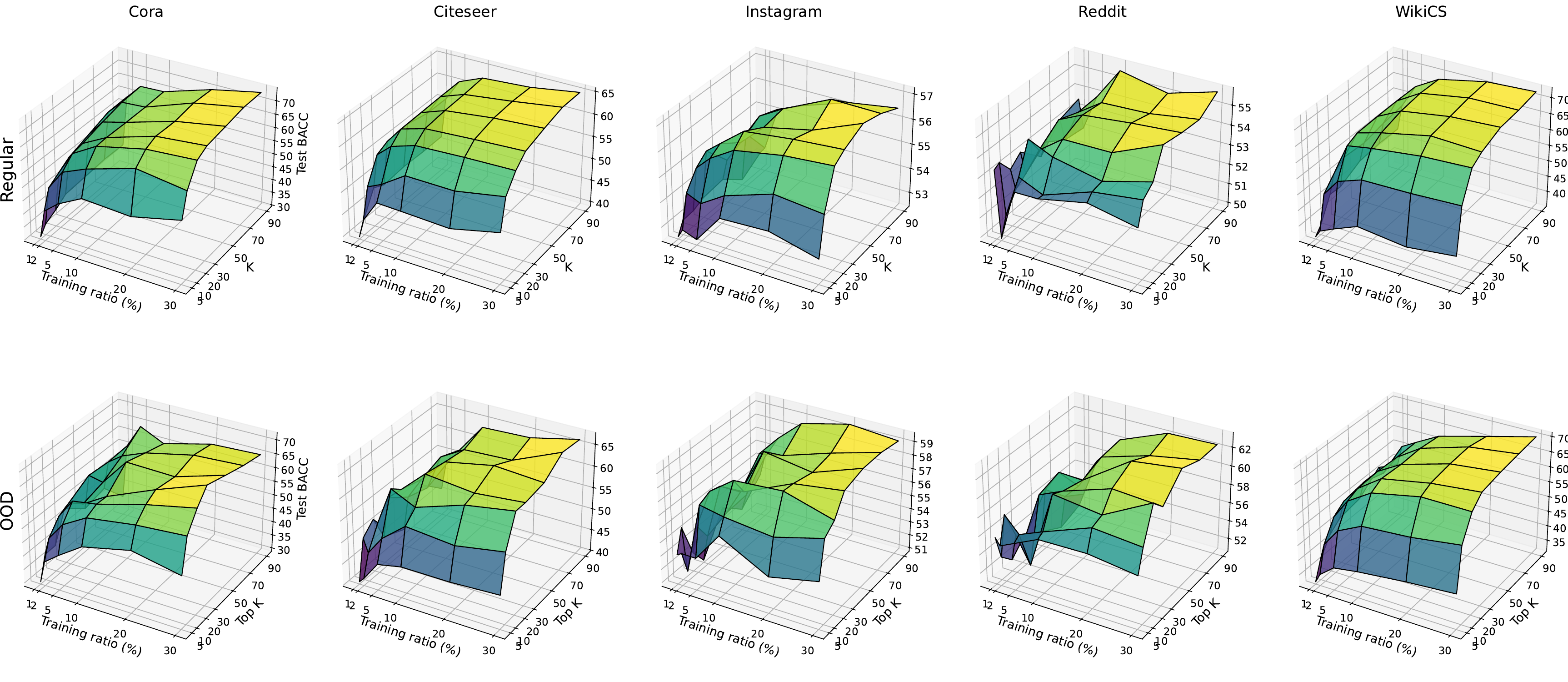}
  \caption{Performance of \model{} across different concept sizes ($K$) and training ratios (\%) on regular splits and OOD splits.} 
  \label{fig:topk}
\end{figure*}
% \yue{can we switch the order and introduce OOD and perturbations first?}
\textit{(1) \model{} improves the model generalizability in OOD data. }
We evaluate \model{} under the \textit{OOD setting} across different upsampling ratios. We report the test results in Table~\ref{tab:res_ood} for the upsampling ratio \(\gamma = 5\); the complete results for all upsampling ratios are provided in ~\ref{sec::addtional}. The results show that \model{} not only significantly outperforms all self-explainable graph learning methods, but also consistently surpasses SOTA GNNs. We attribute this to \model{}’s reliance on a constrained set of concepts for prediction, which potentially contains fewer spurious features and thus makes it less susceptible to distribution shifts. \model{} is therefore a strong baseline for improving OOD generalizability in graph learning.

% \yue{table 2 OOD results on history and children are better; remove the classes that have few exmpales in history and children}

\textit{(2) \model{} improves model robustness under data perturbations.}  
We evaluate \model{} under the \textit{adversarial setting} with different perturbation ratios. We only report the test results in Table~\ref{tab:res_pertube} for perturbation ratio \(\rho = 0.3\); full results for all perturbation ratios are provided in Appendix~\ref{sec::addtional}. 
We observe that while most GNNs perform well under clean conditions, their performance degrades significantly when trained on perturbed data, highlighting their vulnerability to poisoning attacks. In contrast, \model{} demonstrates strong robustness against perturbed train data, while maintaining performance comparable to the model trained on clean data. We attribute this robustness to the use of a pretrained graph encoder trained on augmented data from diverse domains.

\textit{(3) \model{} incurs minimal cost in model utility on clean in-distribution data.}  
We evaluate \model{} and baseline methods under the \textit{regular setting}, and report the test BACC scores (averaged over 5 trials) in Table~\ref{tab:res_regular}. On three out of five datasets, \model{} achieves the best performance (in at least one metric) among interpretable GNN methods. Moreover, compared to the overall best-performing model, \model{} delivers competitive results with only small performance gaps, demonstrating its ability to retain high predictive utility while offering interpretability.

\subsection{\model{} as an Explainer} 
\label{sec::info_leakge}

We now examine the quality of the interpretation offered by \model{}. Specifically, we evaluate it in the following terms:
\vspace{-10pt}
\begin{itemize}
  \setlength{\topsep}{0pt}
  \setlength{\itemsep}{0pt}
  \setlength{\parskip}{0pt}
  \setlength{\parsep}{0pt}
  \item \emph{Faithfulness}. Is it meaningful and relevant?
  \item \emph{Necessity} and \emph{sufficiency}. How many concepts actually carry each decision?
  \item \emph{Sparsity}. Is it concise and human-tractable?
\end{itemize}
\vspace{-10pt}

Together, they assess whether the concepts are functionally responsible for the model’s decisions and whether the resulting explanations are concise, faithful, and meaningful.

(1) \textbf{Faithfulness.}
While the quality of the explanation and prediction performance are correlated due to its self-explainable nature, 
there is a concern about information leakage, which can undermine the faithfulness~\cite{Havasi2022, sun2024eliminatinginfo}. When training the label predictor, it might exploit spurious signals from the concept activations produced by the concept predictor rather than relying on the semantics of the concepts. In such cases, even if the concepts are meaningless or unrelated, the model could still achieve high accuracy by assigning higher activation scores to arbitrary concepts that correlate with the label, providing no true explainable value.
 Thus, we want to answer a crucial question:
\textit{Is the accurate prediction provided by \model{} driven by a faithful projection of the input into a semantic space that is genuinely relevant to the label, or is it merely the result of information leakage?}
Inspired by \cite{mahinpei2021spitfall}, since there is no easy way to directly measure information leakage, we assess it indirectly by replacing the retrieved concepts with random ones. Intuitively, \textit{if the model maintains strong performance with meaningless concepts, this indicates that its predictions do not truly rely on the intended concept subspace, but instead exploit alternative pathways, suggesting the presence of information leakage}. To eliminate any real-world semantics, we construct a set of random concepts consisting of arbitrary tokens representing random numbers (e.g., `1', `99', etc.),
We report the performance of \model{} using both retrieved concepts (“GCB”) and random concepts (“GCB-RC”) across different numbers of selected concepts $K$, under \textit{regular} and \textit{OOD} settings in Figure~\ref{fig:rc}. The difference between the two plotted as a gray bar can measure the degree of information leakage.
For the regular split, we observe a general pattern: the performance gap gradually decreases as the concept size increases. Specifically, GCB-RC performs significantly worse with smaller concept sizes but gradually approaches GCB’s performance as the concept size grows. This suggests that when the concept set is large enough, the model may rely more on spurious correlations between concept activation patterns and labels. Conversely, when the concept set is small, the spurious patterns are harder to explore, and the relevance of actual concepts plays a more critical role.
For the OOD split, random concepts fail across all concept sizes, highlighting the inability of random concepts to generalize beyond the training distribution.
These findings indicate that although random concepts can achieve reasonable performance with a sufficiently large concept set under in-distribution data, they fail when the concept set is limited or when distribution shifts occur. This highlights the importance of carefully controlling the size of the concept set: \textit{using a small, curated set of concepts not only enhances the interpretability of the model, but also helps ensure the quality and faithfulness of its explanations.}

(2) \textbf{Necessity and sufficiency.}
A good explanation should be both \emph{necessary} and \emph{sufficient}. Using too many concepts can hurt interpretability and, as discussed in the previous section, make the model more susceptible to information leakage, while using too few may deprive the model of essential information and impair predictive accuracy. We therefore examine the sensitivity of \model{} to the concept size
$K$ under varying training ratios for each dataset. The results are visualized in Figure~\ref{fig:topk}, where the x-axis denotes the training ratio and the y-axis indicates the number of concepts.
Across datasets, we observe a consistent pattern: \textit{performance improves rapidly as 
$K$ increases, but the gains gradually diminish and eventually plateau.} This indicates that a relatively small number of concepts is sufficient to capture most of the predictive signal. In the OOD setting, however, enlarging the concept set can even degrade performance, particularly when training ratio is small. Having too many concepts appears to hinder generalization, likely by introducing spurious or unstable signals.
Overall, it suggest that a moderate concept size offers the best trade-off between expressiveness and robustness. In practice, selecting $K$ in the range of 30–50 achieves strong performance while preserving the necessity and sufficiency of the explanations.
\begin{figure*}[h]
  \centering
    \includegraphics[width=0.9\textwidth]{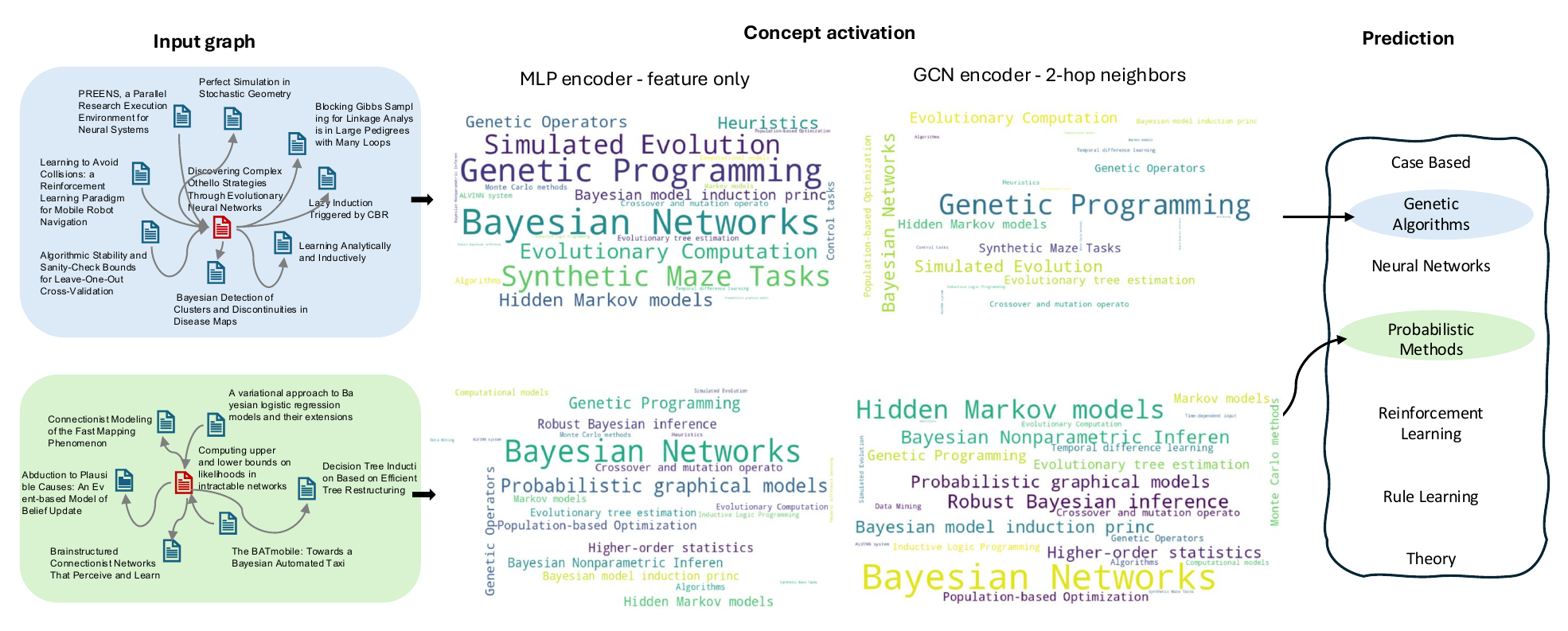}
  \caption{Case study on \texttt{cora}. We visualize the concept activations of two instances produced by both the MLP encoder, which uses only node features, and the GCN encoder, which incorporates 2-hop neighborhood information. %The results show that incorporating graph structure helps the model activate concepts that are more relevant to the class label. 
  }
  \label{fig:case}
\end{figure*}

(3) \textbf{Sparsity.} We study whether \model{} can provide concise and human-tractable explanations via a case study on \texttt{cora} with two sampled instances as shown in Figure~\ref{fig:case}. For each instance, we visualize the concept activations as word clouds, where font size reflects activation scores. We compare activations produced by an MLP encoder (feature-only) and a GCN encoder (with 2-hop neighborhoods) to reveal the role of graph structure. In the upper graph, the MLP highlights \textit{Genetic Programming, Simulated Evolution, and Genetic Operators}, but also activates less relevant concepts such as \textit{Synthetic Maze Tasks} and \textit{Bayesian Networks}. In contrast, the GCN concentrates on the most pertinent evolutionary concepts, indicating that neighborhood information helps sharpen semantic focus. Similarly, for the lower graph, the MLP emphasizes \textit{Bayesian Networks} and \textit{Probabilistic Graphical Models} yet also activates \textit{Genetic Programming}, whereas the GCN primarily activates highly relevant concepts such as \textit{Hidden Markov Models} and \textit{Bayesian Inference}. These comparisons demonstrate that incorporating graph structure yields more accurate concept activations. We also emphasize that the explanation is not determined by only a few top-activated concepts, but by the entire activation landscape across all concepts. When the  concept activation is properly visualized—such as through a word cloud—it enables users to gain intuitive insights into the model’s decision process that go beyond what can be conveyed by natural language alone.

% We investigate how \model{} explains model predictions through a case study. For each dataset, we sample test instances that are predicted to belong to each class and examine the corresponding concept activation vectors. This allows us to analyze which concepts are (in)active in relation to the predicted labels. Specifically, we visualize the average concept activations of 10 sampled instances per class across all selected concepts (\(K=30\)) as word clouds. Figure~\ref{fig:word_cloud} presents the word clouds for three classes from \texttt{WikiCS}, where concepts like ``Live USB,'' ``Baikal CPU,'' and ``Encryption'' are prominently activated for three different predicted classes. We also use Sankey diagrams to visualize the concept activations for three classes in Figure \ref{fig:sankey_wikics}, showing how the model distinguishes between different classes. The complete set of word clouds and Sankey diagrams for all datasets is provided in ~\ref{append:interpretability}.
% They illustrate that the concept activations provide an intuitive and class-discriminative explanation of the model's decision-making process.

\begin{figure*}[h]
  \centering
    \includegraphics[width=\textwidth]{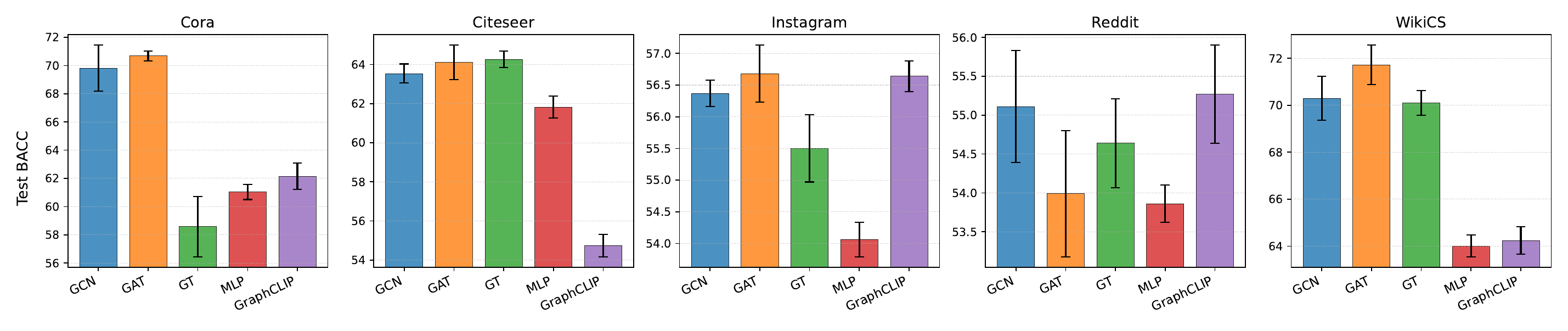}
  \caption{Performance of \model{} variations using different graph encoders.}
  \label{fig:encoder}
\end{figure*}
\subsection{Graph encoders.} 
We study how different graph-text alignment models affect performance. First, we compare versions of GCB using different graph encoders: GCN (default), GAT, and Graph Transformer (GT). We also explore the effect of removing the graph structure by replacing the graph encoder with an MLP for decoding the concept map. Additionally, we evaluate a pretrained graph foundation model, GraphCLIP, which includes both graph and text encoders for graph-text alignment. All results are shown in Figure~\ref{fig:encoder}.
We observe that GCN consistently performs well across all datasets compared to GAT and GT, suggesting that a simpler architecture may be more stable when pretraining data is limited. The model’s performance drops significantly when using the MLP encoder, \textit{highlighting the importance of leveraging graph structure for mapping input graphs into the concept space.} GraphCLIP performs slightly better on \texttt{Instagram} and \texttt{Reddit} but considerably worse on the other three datasets.
One potential reason is that GraphCLIP aligns graphs to free-form summaries that contain noisy information, which can lead to inaccurate mappings between graphs and concepts. Moreover, since GraphCLIP jointly trains both the graph and text encoders and may cause overfitting, especially when the training corpus is small or domain-specific. This limitation could explain why GraphCLIP performs well on \texttt{Instagram} and \texttt{Reddit}—social networks that likely share overlapping topics with its training corpus—but poorly on \texttt{Cora}, \texttt{Citeseer}, and \texttt{WikiCS}.

% \label{sec::interpretability}

\section{Further Discussion \& Conclusion}

\textit{\model{} vs. LLM-as-predictor methods.} While some LLM-as-predictor/explainer approaches~\cite{wang2024llmszeroshotgraphlearners, chen2024llagalargelanguagegraph, tang2024graphgptgraphinstructiontuning, pan2025graphnarratorgeneratingtextualexplanations} can produce predictions accompanied by natural language explanations that may appear informative, they are fundamentally different. (1) Their explanations are inherently \textit{post-hoc}: the generated text is not guaranteed to faithfully reflect the actual reasoning process, and how these explanations are produced remains another black box. In contrast, \model{} makes predictions directly based on the semantics of concepts, ensuring that explanations are faithful and aligned with the model’s prediction. (2) \model{} requires access to LLMs only during training. At inference time, no LLM queries are needed. In comparison, LLM-as-predictor methods rely on querying LLMs for each prediction, which incurs substantial computational and monetary costs.

\textit{\model{}’s generalizability across differently graph types.} 
The performance of \model{} largely depends on the quality of the proposed concept space and the effectiveness of the graph-concept alignment model—both of which rely on LLMs for semantic  understanding and reasoning over graph instances. To date, LLM-based graph reasoning has primarily focused on text-attributed graphs, which motivates our choice of such graphs as the starting point for exploring \model{}. Nevertheless, \model{} holds strong potential for broader applicability to diverse graph types, such as molecular and biomedical graphs, provided that suitable LLM-driven interfaces~\cite{wang2025bridgingmoleculargraphslarge, lee2025molllmgeneralistmolecularllm, bran2023chemcrowaugmentinglargelanguagemodels} are available to bridge domain-specific graph structures with high-level concepts. We plan to explore this direction as future work.

% Possible ideas for discussion section: (1) what is the key ingredient for a successful alignment of the graph and the concept spaces? e.g., the contrastive loss? (2) We can also discuss the nature of the graph reasoning tasks: what capabilities are needed to solve these tasks well? E.g., the noun concepts, the relations between the objects, etc. (3) Argue for the CBM avenue compared to the LLM avenue. E.g., CBM is transparent by design: LLMs capture much information but it's unclear how they did that. 

\textbf{Conclusion}
We present \model{} as a novel solution for self-explainable text-attributed graph learning. \model{} maps graph inputs into a human-interpretable concept space, where each concept is in natural language and carries semantics. Predictions are then made directly based on these concepts. We conduct extensive experiments and case studies on five real-world datasets from diverse domains to demonstrate the effectiveness of \model{} both as a predictor and as an explainer. 


\section*{Impact Statement}
This paper presents work whose goal is to advance the field of graph learning. There are many potential societal consequences of our work, none of which we feel must be specifically highlighted here.
\bibliography{example_paper}

@inproceedings{kipf2017GCN,
title={Semi-Supervised Classification with Graph Convolutional Networks},
author={Thomas N. Kipf and Max Welling},
booktitle={International Conference on Learning Representations},
year={2017},
}

@inproceedings{veličković2018GAT,
title={Graph Attention Networks},
author={Petar Veličković and Guillem Cucurull and Arantxa Casanova and Adriana Romero and Pietro Liò and Yoshua Bengio},
booktitle={International Conference on Learning Representations},
year={2018},
}

@inproceedings{hamilton2018SAGE,
author = {Hamilton, William L. and Ying, Rex and Leskovec, Jure},
title = {Inductive representation learning on large graphs},
year = {2017},
isbn = {9781510860964},
publisher = {Curran Associates Inc.},
address = {Red Hook, NY, USA},
booktitle = {Proceedings of the 31st International Conference on Neural Information Processing Systems},
pages = {1025–1035},
numpages = {11},
location = {Long Beach, California, USA},
series = {NIPS'17}
}

@inproceedings{yun2020GT,
 author = {Yun, Seongjun and Jeong, Minbyul and Kim, Raehyun and Kang, Jaewoo and Kim, Hyunwoo J},
 booktitle = {Advances in Neural Information Processing Systems},
 __editor = {H. Wallach and H. Larochelle and A. Beygelzimer and F. d\textquotesingle Alch\'{e}-Buc and E. Fox and R. Garnett},
 pages = {},
 publisher = {Curran Associates, Inc.},
 title = {Graph Transformer Networks},
 volume = {32},
 year = {2019}
}

@misc{kakkad2023-survey,
      title={A Survey on Explainability of Graph Neural Networks}, 
      author={Jaykumar Kakkad and Jaspal Jannu and Kartik Sharma and Charu Aggarwal and Sourav Medya},
      year={2023},
      eprint={2306.01958},
      archivePrefix={arXiv},
      primaryClass={cs.LG},
}

@InProceedings{yu2022-GIB,
    author    = {Yu, Junchi and Cao, Jie and He, Ran},
    title     = {Improving Subgraph Recognition With Variational Graph Information Bottleneck},
    booktitle = {Proceedings of the IEEE/CVF Conference on Computer Vision and Pattern Recognition (CVPR)},
    month     = {June},
    year      = {2022},
    pages     = {19396-19405}
}

@inproceedings{yu2020-GIB,
title={Graph Information Bottleneck for Subgraph Recognition},
author={Junchi Yu and Tingyang Xu and Yu Rong and Yatao Bian and Junzhou Huang and Ran He},
booktitle={International Conference on Learning Representations},
year={2021},
}

@InProceedings{miao2022-GIB,
  title = 	 {Interpretable and Generalizable Graph Learning via Stochastic Attention Mechanism},
  author =       {Miao, Siqi and Liu, Mia and Li, Pan},
  booktitle = 	 {Proceedings of the 39th International Conference on Machine Learning},
  pages = 	 {15524--15543},
  year = 	 {2022},
  __editor = 	 {Chaudhuri, Kamalika and Jegelka, Stefanie and Song, Le and Szepesvari, Csaba and Niu, Gang and Sabato, Sivan},
  volume = 	 {162},
  series = 	 {Proceedings of Machine Learning Research},
  month = 	 {17--23 Jul},
  publisher =    {PMLR},
  pdf = 	 {https://proceedings.mlr.press/v162/miao22a/miao22a.pdf},
  abstract = 	 {Interpretable graph learning is in need as many scientific applications depend on learning models to collect insights from graph-structured data. Previous works mostly focused on using post-hoc approaches to interpret pre-trained models (graph neural networks in particular). They argue against inherently interpretable models because the good interpretability of these models is often at the cost of their prediction accuracy. However, those post-hoc methods often fail to provide stable interpretation and may extract features that are spuriously correlated with the task. In this work, we address these issues by proposing Graph Stochastic Attention (GSAT). Derived from the information bottleneck principle, GSAT injects stochasticity to the attention weights to block the information from task-irrelevant graph components while learning stochasticity-reduced attention to select task-relevant subgraphs for interpretation. The selected subgraphs provably do not contain patterns that are spuriously correlated with the task under some assumptions. Extensive experiments on eight datasets show that GSAT outperforms the state-of-the-art methods by up to 20% in interpretation AUC and 5% in prediction accuracy. Our code is available at https://github.com/Graph-COM/GSAT.}
}

@inproceedings{miao2023-GIB,
title = {Interpretable Geometric Deep Learning via Learnable Randomness Injection}, 
abstractNote = {}, 
booktitle = {International Conference on Learning Representations}, 
author = {Siqi Miao and Yunan Luo and Mia Liu and Pan Li}, 
year={2023}
}

@inproceedings{wu2020-GIB,
 author = {Wu, Tailin and Ren, Hongyu and Li, Pan and Leskovec, Jure},
 booktitle = {Advances in Neural Information Processing Systems},
 __editor = {H. Larochelle and M. Ranzato and R. Hadsell and M.F. Balcan and H. Lin},
 pages = {20437--20448},
 publisher = {Curran Associates, Inc.},
 title = {Graph Information Bottleneck},
 volume = {33},
 year = {2020}
}

@inproceedings{wu2022-DIR,
title={Discovering Invariant Rationales for Graph Neural Networks},
author={Yingxin Wu and Xiang Wang and An Zhang and Xiangnan He and Tat-Seng Chua},
booktitle={International Conference on Learning Representations},
year={2022},
}

@inproceedings{dai2021-SEGNN,
author = {Dai, Enyan and Wang, Suhang},
title = {Towards Self-Explainable Graph Neural Network},
year = {2021},
isbn = {9781450384469},
publisher = {Association for Computing Machinery},
address = {New York, NY, USA},
doi = {10.1145/3459637.3482306},
booktitle = {Proceedings of the 30th ACM International Conference on Information \& Knowledge Management},
pages = {302–311},
numpages = {10},
location = {Virtual Event, Queensland, Australia},
series = {CIKM'21}
}

@article{feng2022-KERGNN, 
title={{KerGNNs: Interpretable Graph Neural Networks with Graph Kernels}}, 
volume={36}, 
DOI={10.1609/aaai.v36i6.20615}, 
abstractNote={Graph kernels are historically the most widely-used technique for graph classification tasks. However, these methods suffer from limited performance because of the hand-crafted combinatorial features of graphs. In recent years, graph neural networks (GNNs) have become the state-of-the-art method in downstream graph-related tasks due to their superior performance. Most GNNs are based on Message Passing Neural Network (MPNN) frameworks. However, recent studies show that MPNNs can not exceed the power of the Weisfeiler-Lehman (WL) algorithm in graph isomorphism test. To address the limitations of existing graph kernel and GNN methods, in this paper, we propose a novel GNN framework, termed Kernel Graph Neural Networks (KerGNNs), which integrates graph kernels into the message passing process of GNNs. Inspired by convolution filters in convolutional neural networks (CNNs), KerGNNs adopt trainable hidden graphs as graph filters which are combined with subgraphs to update node embeddings using graph kernels. In addition, we show that MPNNs can be viewed as special cases of KerGNNs. We apply KerGNNs to multiple graph-related tasks and use cross-validation to make fair comparisons with benchmarks. We show that our method achieves competitive performance compared with existing state-of-the-art methods, demonstrating the potential to increase the representation ability of GNNs. We also show that the trained graph filters in KerGNNs can reveal the local graph structures of the dataset, which significantly improves the model interpretability compared with conventional GNN models.}, 
number={6}, 
journal={Proceedings of the AAAI Conference on Artificial Intelligence}, 
author={Feng, Aosong and You, Chenyu and Wang, Shiqiang and Tassiulas, Leandros}, 
year={2022}, 
month={Jun.}, 
pages={6614-6622} 
}

@InProceedings{koh2020conceptbottleneckmodels,
  title = 	 {Concept Bottleneck Models},
  author =       {Koh, Pang Wei and Nguyen, Thao and Tang, Yew Siang and Mussmann, Stephen and Pierson, Emma and Kim, Been and Liang, Percy},
  booktitle = 	 {Proceedings of the 37th International Conference on Machine Learning},
  pages = 	 {5338--5348},
  year = 	 {2020},
  __editor = 	 {III, Hal Daumé and Singh, Aarti},
  volume = 	 {119},
  series = 	 {Proceedings of Machine Learning Research},
  month = 	 {13--18 Jul},
  publisher =    {PMLR},
  pdf = 	 {http://proceedings.mlr.press/v119/koh20a/koh20a.pdf},
  abstract = 	 {We seek to learn models that we can interact with using high-level concepts: if the model did not think there was a bone spur in the x-ray, would it still predict severe arthritis? State-of-the-art models today do not typically support the manipulation of concepts like "the existence of bone spurs", as they are trained end-to-end to go directly from raw input (e.g., pixels) to output (e.g., arthritis severity). We revisit the classic idea of first predicting concepts that are provided at training time, and then using these concepts to predict the label. By construction, we can intervene on these concept bottleneck models by editing their predicted concept values and propagating these changes to the final prediction. On x-ray grading and bird identification, concept bottleneck models achieve competitive accuracy with standard end-to-end models, while enabling interpretation in terms of high-level clinical concepts ("bone spurs") or bird attributes ("wing color"). These models also allow for richer human-model interaction: accuracy improves significantly if we can correct model mistakes on concepts at test time.}
}

@inproceedings{oikarinen2023labelfreeconceptbottleneckmodels,
  title={Label-free Concept Bottleneck Models},
  author={Oikarinen, Tuomas and Das, Subhro and Nguyen, Lam M and Weng, Tsui-Wei},
  booktitle={International Conference on Learning Representations},
  year={2023}
}

@misc{
xu2025graph,
title={Graph Concept Bottleneck Models},
author={Haotian Xu and Tsui-Wei Weng and Lam M. Nguyen and Tengfei Ma},
year={2025},
}

@InProceedings{shin2023closerlookCBM,
  title = 	 {A Closer Look at the Intervention Procedure of Concept Bottleneck Models},
  author =       {Shin, Sungbin and Jo, Yohan and Ahn, Sungsoo and Lee, Namhoon},
  booktitle = 	 {Proceedings of the 40th International Conference on Machine Learning},
  pages = 	 {31504--31520},
  year = 	 {2023},
  __editor = 	 {Krause, Andreas and Brunskill, Emma and Cho, Kyunghyun and Engelhardt, Barbara and Sabato, Sivan and Scarlett, Jonathan},
  volume = 	 {202},
  series = 	 {Proceedings of Machine Learning Research},
  month = 	 {23--29 Jul},
  publisher =    {PMLR},
  pdf = 	 {https://proceedings.mlr.press/v202/shin23a/shin23a.pdf},
  abstract = 	 {Concept bottleneck models (CBMs) are a class of interpretable neural network models that predict the target response of a given input based on its high-level concepts. Unlike the standard end-to-end models, CBMs enable domain experts to intervene on the predicted concepts and rectify any mistakes at test time, so that more accurate task predictions can be made at the end. While such intervenability provides a powerful avenue of control, many aspects of the intervention procedure remain rather unexplored. In this work, we develop various ways of selecting intervening concepts to improve the intervention effectiveness and conduct an array of in-depth analyses as to how they evolve under different circumstances. Specifically, we find that an informed intervention strategy can reduce the task error more than ten times compared to the current baseline under the same amount of intervention counts in realistic settings, and yet, this can vary quite significantly when taking into account different intervention granularity. We verify our findings through comprehensive evaluations, not only on the standard real datasets, but also on synthetic datasets that we generate based on a set of different causal graphs. We further discover some major pitfalls of the current practices which, without a proper addressing, raise concerns on reliability and fairness of the intervention procedure.}
}

@INPROCEEDINGS{shang2024incrementalCBM,
  author={Shang, Chenming and Zhou, Shiji and Zhang, Hengyuan and Ni, Xinzhe and Yang, Yujiu and Wang, Yuwang},
  booktitle={2024 IEEE/CVF Conference on Computer Vision and Pattern Recognition (CVPR)}, 
  title={Incremental Residual Concept Bottleneck Models}, 
  year={2024},
  volume={},
  number={},
  pages={11030-11040},
  doi={10.1109/CVPR52733.2024.01049}
}

@inproceedings{yuksekgonul2023posthocCBM,
title={Post-hoc Concept Bottleneck Models},
author={Mert Yuksekgonul and Maggie Wang and James Zou},
booktitle={The Eleventh International Conference on Learning Representations},
year={2023},
}

@inproceedings{kim2023probabilisticCBM,
author = {Kim, Eunji and Jung, Dahuin and Park, Sangha and Kim, Siwon and Yoon, Sungroh},
title = {Probabilistic concept bottleneck models},
year = {2023},
publisher = {JMLR.org},
booktitle = {Proceedings of the 40th International Conference on Machine Learning},
articleno = {677},
numpages = {20},
location = {Honolulu, Hawaii, USA},
series = {ICML'23}
}

@InProceedings{radford2021learningtransferablevisualmodels,
  title = 	 {Learning Transferable Visual Models From Natural Language Supervision},
  author =       {Radford, Alec and Kim, Jong Wook and Hallacy, Chris and Ramesh, Aditya and Goh, Gabriel and Agarwal, Sandhini and Sastry, Girish and Askell, Amanda and Mishkin, Pamela and Clark, Jack and Krueger, Gretchen and Sutskever, Ilya},
  booktitle = 	 {Proceedings of the 38th International Conference on Machine Learning},
  pages = 	 {8748--8763},
  year = 	 {2021},
  __editor = 	 {Meila, Marina and Zhang, Tong},
  volume = 	 {139},
  series = 	 {Proceedings of Machine Learning Research},
  month = 	 {18--24 Jul},
  publisher =    {PMLR},
  pdf = 	 {http://proceedings.mlr.press/v139/radford21a/radford21a.pdf},
  abstract = 	 {State-of-the-art computer vision systems are trained to predict a fixed set of predetermined object categories. This restricted form of supervision limits their generality and usability since additional labeled data is needed to specify any other visual concept. Learning directly from raw text about images is a promising alternative which leverages a much broader source of supervision. We demonstrate that the simple pre-training task of predicting which caption goes with which image is an efficient and scalable way to learn SOTA image representations from scratch on a dataset of 400 million (image, text) pairs collected from the internet. After pre-training, natural language is used to reference learned visual concepts (or describe new ones) enabling zero-shot transfer of the model to downstream tasks. We study the performance of this approach by benchmarking on over 30 different existing computer vision datasets, spanning tasks such as OCR, action recognition in videos, geo-localization, and many types of fine-grained object classification. The model transfers non-trivially to most tasks and is often competitive with a fully supervised baseline without the need for any dataset specific training. For instance, we match the accuracy of the original ResNet-50 on ImageNet zero-shot without needing to use any of the 1.28 million training examples it was trained on.}
}

@inproceedings{zhu2025GraphCLIP,
author = {Zhu, Yun and Shi, Haizhou and Wang, Xiaotang and Liu, Yongchao and Wang, Yaoke and Peng, Boci and Hong, Chuntao and Tang, Siliang},
title = {{GraphCLIP: Enhancing Transferability in Graph Foundation Models for Text-Attributed Graphs}},
year = {2025},
isbn = {9798400712746},
publisher = {Association for Computing Machinery},
address = {New York, NY, USA},
doi = {10.1145/3696410.3714801},
booktitle = {Proceedings of the ACM on Web Conference 2025},
pages = {2183–2197},
numpages = {15},
location = {Sydney NSW, Australia},
series = {WWW '25}
}

@InProceedings{han2024decaf,
  title = 	 {{DeCaf: A Causal Decoupling Framework for OOD Generalization on Node Classification}},
  author =       {Han, Xiaoxue and Rangwala, Huzefa and Ning, Yue},
  booktitle = 	 {Proceedings of The 28th International Conference on Artificial Intelligence and Statistics},
  pages = 	 {2332--2340},
  year = 	 {2025},
  __editor = 	 {Li, Yingzhen and Mandt, Stephan and Agrawal, Shipra and Khan, Emtiyaz},
  volume = 	 {258},
  series = 	 {Proceedings of Machine Learning Research},
  month = 	 {03--05 May},
  publisher =    {PMLR},
  pdf = 	 {https://raw.githubusercontent.com/mlresearch/v258/main/assets/han25b/han25b.pdf},
  abstract = 	 {Graph Neural Networks (GNNs) are susceptible to distribution shifts, creating vulnerability and security issues in critical domains. There is a pressing need to enhance the generalizability of GNNs on out-of-distribution (OOD) test data. Existing methods that target learning an invariant (feature, structure)-label mapping often depend on oversimplified assumptions about the data generation process, which do not adequately reflect the actual dynamics of distribution shifts in graphs. In this paper, we introduce a more realistic graph data generation model using Structural Causal Models (SCMs), allowing us to redefine distribution shifts by pinpointing their origins within the generation process. Building on this, we propose a casual decoupling framework, DeCaf, that independently learns unbiased feature-label and structure-label mappings. We provide a detailed theoretical framework that shows how our approach can effectively mitigate the impact of various distribution shifts. We evaluate DeCaf across both real-world and synthetic datasets that demonstrate different patterns of shifts, confirming its efficacy in enhancing the generalizability of GNNs. Our code is available at: \url{https://github.com/hanxiaoxue114/DeCaf-GraphOOD.}}
}

@inproceedings{reimers-2019-sentence-bert,
    title = {{Sentence-BERT: Sentence Embeddings using Siamese BERT-Networks}},
    author = "Reimers, Nils and Gurevych, Iryna",
    booktitle = "Proceedings of the 2019 Conference on Empirical Methods in Natural Language Processing",
    month = "11",
    year = "2019",
    publisher = "Association for Computational Linguistics",
}

@inproceedings{brown2020gpt3,
 author = {Brown, Tom and Mann, Benjamin and Ryder, Nick and Subbiah, Melanie and Kaplan, Jared D and Dhariwal, Prafulla and Neelakantan, Arvind and Shyam, Pranav and Sastry, Girish and Askell, Amanda and Agarwal, Sandhini and Herbert-Voss, Ariel and Krueger, Gretchen and Henighan, Tom and Child, Rewon and Ramesh, Aditya and Ziegler, Daniel and Wu, Jeffrey and Winter, Clemens and Hesse, Chris and Chen, Mark and Sigler, Eric and Litwin, Mateusz and Gray, Scott and Chess, Benjamin and Clark, Jack and Berner, Christopher and McCandlish, Sam and Radford, Alec and Sutskever, Ilya and Amodei, Dario},
 booktitle = {Advances in Neural Information Processing Systems},
 editor = {H. Larochelle and M. Ranzato and R. Hadsell and M.F. Balcan and H. Lin},
 pages = {1877--1901},
 publisher = {Curran Associates, Inc.},
 title = {Language Models are Few-Shot Learners},
 volume = {33},
 year = {2020}
}

@article{oord2018representation,
  title={Representation Learning with Contrastive Predictive Coding},
  author={van den Oord, Aaron and Li, Yazhe and Vinyals, Oriol},
  journal={arXiv preprint arXiv:1807.03748},
  year={2018}
}

@inproceedings{lee2024LLMdomain,
    title = "Towards Understanding Counseling Conversations: Domain Knowledge and Large Language Models",
    author = "Lee, Younghun  and
      Goldwasser, Dan  and
      Reese, Laura Schwab",
    __editor = "Graham, Yvette  and
      Purver, Matthew",
    booktitle = "Findings of the Association for Computational Linguistics: EACL 2024",
    month = mar,
    year = "2024",
    address = "St. Julian{'}s, Malta",
    publisher = "Association for Computational Linguistics",
    pages = "2032--2047",
    abstract = "Understanding the dynamics of counseling conversations is an important task, yet it is a challenging NLP problem regardless of the recent advance of Transformer-based pre-trained language models. This paper proposes a systematic approach to examine the efficacy of domain knowledge and large language models (LLMs) in better representing conversations between a crisis counselor and a help seeker. We empirically show that state-of-the-art language models such as Transformer-based models and GPT models fail to predict the conversation outcome. To provide richer context to conversations, we incorporate human-annotated domain knowledge and LLM-generated features; simple integration of domain knowledge and LLM features improves the model performance by approximately 15{\%}. We argue that both domain knowledge and LLM-generated features can be exploited to better characterize counseling conversations when they are used as an additional context to conversations."
}

@article{lee2024LLMreasoning,
author = {Lee, Seungpil and Sim, Woochang and Shin, Donghyeon and Seo, Wongyu and Park, Jiwon and Lee, Seokki and Hwang, Sanha and Kim, Sejin and Kim, Sundong},
title = {Reasoning Abilities of Large Language Models: In-Depth Analysis on the Abstraction and Reasoning Corpus},
year = {2025},
publisher = {Association for Computing Machinery},
address = {New York, NY, USA},
issn = {2157-6904},
doi = {10.1145/3712701},
note = {Just Accepted},
journal = {ACM Trans. Intell. Syst. Technol.},
month = jan
}

@misc{zhang2024llmreasoning,
      title={{Chain-of-Knowledge: Integrating Knowledge Reasoning into Large Language Models by Learning from Knowledge Graphs}}, 
      author={Yifei Zhang and Xintao Wang and Jiaqing Liang and Sirui Xia and Lida Chen and Yanghua Xiao},
      year={2024},
      eprint={2407.00653},
      archivePrefix={arXiv},
      primaryClass={cs.CL},
}

@inproceedings{wang2024llmszeroshotgraphlearners,
 author = {Wang, Duo and Zuo, Yuan and Li, Fengzhi and Wu, Junjie},
 booktitle = {Advances in Neural Information Processing Systems},
 __editor = {A. Globerson and L. Mackey and D. Belgrave and A. Fan and U. Paquet and J. Tomczak and C. Zhang},
 pages = {5950--5973},
 publisher = {Curran Associates, Inc.},
 title = {LLMs as Zero-shot Graph Learners: Alignment of GNN Representations with LLM Token Embeddings},
 volume = {37},
 year = {2024}
}

@InProceedings{chen2024llagalargelanguagegraph,
  title = 	 {{LL}a{GA}: Large Language and Graph Assistant},
  author =       {Chen, Runjin and Zhao, Tong and Jaiswal, Ajay Kumar and Shah, Neil and Wang, Zhangyang},
  booktitle = 	 {Proceedings of the 41st International Conference on Machine Learning},
  pages = 	 {7809--7823},
  year = 	 {2024},
  __editor = 	 {Salakhutdinov, Ruslan and Kolter, Zico and Heller, Katherine and Weller, Adrian and Oliver, Nuria and Scarlett, Jonathan and Berkenkamp, Felix},
  volume = 	 {235},
  series = 	 {Proceedings of Machine Learning Research},
  month = 	 {21--27 Jul},
  publisher =    {PMLR},
  pdf = 	 {https://raw.githubusercontent.com/mlresearch/v235/main/assets/chen24bh/chen24bh.pdf},
  }

@inproceedings{tang2024graphgptgraphinstructiontuning,
author = {Tang, Jiabin and Yang, Yuhao and Wei, Wei and Shi, Lei and Su, Lixin and Cheng, Suqi and Yin, Dawei and Huang, Chao},
title = {{GraphGPT: Graph Instruction Tuning for Large Language Models}},
year = {2024},
isbn = {9798400704314},
publisher = {Association for Computing Machinery},
address = {New York, NY, USA},
doi = {10.1145/3626772.3657775},
booktitle = {Proceedings of the 47th International ACM SIGIR Conference on Research and Development in Information Retrieval},
pages = {491–500},
numpages = {10},
location = {Washington DC, USA},
series = {SIGIR'24}
}

@inproceedings{alemi2019IB,
title={Deep Variational Information Bottleneck},
author={Alexander A. Alemi and Ian Fischer and Joshua V. Dillon and Kevin Murphy},
booktitle={International Conference on Learning Representations},
year={2017},
}

@inproceedings{huang2024Can,
  title={{Can GNN be Good Adapter for LLMs?}},
  author={Huang, Xuanwen and Han, Kaiqiao and Yang, Yang and Bao, Dezheng and Tao, Quanjin and Chai, Ziwei and Zhu, Qi},
  booktitle={Proceedings of the ACM Web Conference 2024},
  pages={893--904},
  year={2024}
}

@inproceedings{xu2018how,
title={How Powerful are Graph Neural Networks?},
author={Keyulu Xu and Weihua Hu and Jure Leskovec and Stefanie Jegelka},
booktitle={International Conference on Learning Representations},
year={2019},
}

@article{mernyei2020wiki,
  title={{Wiki-CS: A Wikipedia-based benchmark for graph neural networks}},
  author={Mernyei, P{\'e}ter and Cangea, C{\u{a}}t{\u{a}}lina},
  journal={arXiv preprint arXiv:2007.02901},
  year={2020}
}

@article{Sen2008cora, 
title={Collective Classification in Network Data}, volume={29},  DOI={10.1609/aimag.v29i3.2157}, abstractNote={Many real-world applications produce networked data such as the world-wide web (hypertext documents connected via hyperlinks), social networks (for example, people connected by friendship links), communication networks (computers connected via communication links) and biological networks (for example, protein interaction networks). A recent focus in machine learning research has been to extend traditional machine learning classification techniques to classify nodes in such networks. In this article, we provide a brief introduction to this area of research and how it has progressed during the past decade. We introduce four of the most widely used inference algorithms for classifying networked data and empirically compare them on both synthetic and real-world data.}, number={3}, journal={AI Magazine}, author={Sen, Prithviraj and Namata, Galileo and Bilgic, Mustafa and Getoor, Lise and Galligher, Brian and Eliassi-Rad, Tina}, year={2008}, month={Sep.}, pages={93} }

@inproceedings{Yan2023TAG,
 author = {Yan, Hao and Li, Chaozhuo and Long, Ruosong and Yan, Chao and Zhao, Jianan and Zhuang, Wenwen and Yin, Jun and Zhang, Peiyan and Han, Weihao and Sun, Hao and Deng, Weiwei and Zhang, Qi and Sun, Lichao and Xie, Xing and Wang, Senzhang},
 booktitle = {Advances in Neural Information Processing Systems},
 __editor = {A. Oh and T. Naumann and A. Globerson and K. Saenko and M. Hardt and S. Levine},
 pages = {17238--17264},
 publisher = {Curran Associates, Inc.},
 title = {A Comprehensive Study on Text-attributed Graphs: Benchmarking and Rethinking},
 volume = {36},
 year = {2023}
}

@article{wang2025bridgingmoleculargraphslarge, 
title={Bridging Molecular Graphs and Large Language Models}, 
volume={39}, 
DOI={10.1609/aaai.v39i20.35422}, 
abstractNote={While Large Language Models (LLMs) have shown exceptional generalization capabilities, their ability to process graph data, such as molecular structures, remains limited. To bridge this gap, this paper proposes Graph2Token, an efficient solution that aligns graph tokens to LLM tokens. The key idea is to represent a graph token with the LLM token vocabulary, without fine-tuning the LLM backbone. To achieve this goal, we first construct a molecule-text paired dataset from multi-sources, including CHEBI and HMDB, to train a graph structure encoder, which reduces the distance between graphs and texts representations in the feature space. Then, we propose a novel alignment strategy that associates a graph token with LLM tokens. To further unleash the potential of LLMs, we collect molecular IUPAC name identifiers, which are incorporated into the LLM prompts. By aligning molecular graphs as special tokens, we can activate LLMs’ generalization ability to molecular few-shot learning. Extensive experiments on molecular classification and regression tasks demonstrate the effectiveness of our proposed Graph2Token.}, 
number={20}, 
journal={Proceedings of the AAAI Conference on Artificial Intelligence}, 
author={Wang, Runze and Yang, Mingqi and Shen, Yanming}, 
year={2025}, 
month={Apr.}, 
pages={21234-21242} 
}

@misc{lee2025molllmgeneralistmolecularllm,
      title={Mol-LLM: Generalist Molecular LLM with Improved Graph Utilization}, 
      author={Chanhui Lee and Yuheon Song and YongJun Jeong and Hanbum Ko and Rodrigo Hormazabal and Sehui Han and Kyunghoon Bae and Sungbin Lim and Sungwoong Kim},
      year={2025},
      eprint={2502.02810},
      archivePrefix={arXiv},
      primaryClass={cs.LG},
}

@misc{bran2023chemcrowaugmentinglargelanguagemodels,
      title={ChemCrow: Augmenting large-language models with chemistry tools}, 
      author={Andres M Bran and Sam Cox and Oliver Schilter and Carlo Baldassari and Andrew D White and Philippe Schwaller},
      year={2023},
      eprint={2304.05376},
      archivePrefix={arXiv},
      primaryClass={physics.chem-ph},
}

@misc{han2023mlpinitembarrassinglysimplegnn,
      title={MLPInit: Embarrassingly Simple GNN Training Acceleration with MLP Initialization}, 
      author={Xiaotian Han and Tong Zhao and Yozen Liu and Xia Hu and Neil Shah},
      year={2023},
      eprint={2210.00102},
      archivePrefix={arXiv},
      primaryClass={cs.LG},
}

@misc{mahinpei2021spitfall,
      title={Promises and Pitfalls of Black-Box Concept Learning Models}, 
      author={Anita Mahinpei and Justin Clark and Isaac Lage and Finale Doshi-Velez and Weiwei Pan},
      year={2021},
      eprint={2106.13314},
      archivePrefix={arXiv},
      primaryClass={cs.LG},
}

@misc{sun2024eliminatinginfo,
      title={Eliminating Information Leakage in Hard Concept Bottleneck Models with Supervised, Hierarchical Concept Learning}, 
      author={Ao Sun and Yuanyuan Yuan and Pingchuan Ma and Shuai Wang},
      year={2024},
      eprint={2402.05945},
      archivePrefix={arXiv},
      primaryClass={cs.LG},
}

@inproceedings{Havasi2022,
 author = {Havasi, Marton and Parbhoo, Sonali and Doshi-Velez, Finale},
 booktitle = {Advances in Neural Information Processing Systems},
 editor = {S. Koyejo and S. Mohamed and A. Agarwal and D. Belgrave and K. Cho and A. Oh},
 pages = {23386--23397},
 publisher = {Curran Associates, Inc.},
 title = {Addressing Leakage in Concept Bottleneck Models},
 volume = {35},
 year = {2022}
}

@inproceedings{
azzolin2025beyond,
title={Beyond Topological Self-Explainable {GNN}s: A Formal Explainability Perspective},
author={Steve Azzolin and SAGAR MALHOTRA and Andrea Passerini and Stefano Teso},
booktitle={Forty-second International Conference on Machine Learning},
year={2025},
}

@inproceedings{
bechler-speicher2024the,
title={The Intelligible and Effective Graph Neural Additive Network},
author={Maya Bechler-Speicher and Amir Globerson and Ran Gilad-Bachrach},
booktitle={The Thirty-eighth Annual Conference on Neural Information Processing Systems},
year={2024},
}

@article{Wang2025prototype,
author = {Dai, Enyan and Wang, Suhang},
title = {Towards Prototype-Based Self-Explainable Graph Neural Network},
year = {2025},
issue_date = {February 2025},
publisher = {Association for Computing Machinery},
address = {New York, NY, USA},
volume = {19},
number = {2},
issn = {1556-4681},
doi = {10.1145/3689647},
journal = {ACM Trans. Knowl. Discov. Data},
month = feb,
articleno = {45},
numpages = {20}
}

@misc{liu2025faithfulclasslevelselfexplainabilitygraph,
      title={Towards Faithful Class-level Self-explainability in Graph Neural Networks by Subgraph Dependencies}, 
      author={Fanzhen Liu and Xiaoxiao Ma and Jian Yang and Alsharif Abuadbba and Kristen Moore and Surya Nepal and Cecile Paris and Quan Z. Sheng and Jia Wu},
      year={2025},
      eprint={2508.11513},
      archivePrefix={arXiv},
      primaryClass={cs.LG},
}

@misc{peng2024fewshotselfexplaininggraphneural,
      title={Towards Few-shot Self-explaining Graph Neural Networks}, 
      author={Jingyu Peng and Qi Liu and Linan Yue and Zaixi Zhang and Kai Zhang and Yunhao Sha},
      year={2024},
      eprint={2408.07340},
      archivePrefix={arXiv},
      primaryClass={cs.LG},
}

@inproceedings{
muller2023graphchef,
title={GraphChef: Learning the Recipe of Your Dataset},
author={Peter M{\"u}ller and Lukas Faber and Karolis Martinkus and Roger Wattenhofer},
booktitle={ICML 3rd Workshop on Interpretable Machine Learning in Healthcare (IMLH) },
year={2023}
}

@misc{sengupta2025xnodeselfexplanationneed,
      title={X-Node: Self-Explanation is All We Need}, 
      author={Prajit Sengupta and Islem Rekik},
      year={2025},
      eprint={2508.10461},
      archivePrefix={arXiv},
      primaryClass={cs.LG},
}

@inproceedings{brandes2000graph,
  title={Graph data format workshop report},
  author={Brandes, Ulrik and Marshall, M Scott and North, Stephen C},
  booktitle={International Symposium on Graph Drawing},
  pages={407--409},
  year={2000},
  organization={Springer}
}

@misc{pan2025graphnarratorgeneratingtextualexplanations,
      title={GraphNarrator: Generating Textual Explanations for Graph Neural Networks}, 
      author={Bo Pan and Zhen Xiong and Guanchen Wu and Zheng Zhang and Yifei Zhang and Liang Zhao},
      year={2025},
      eprint={2410.15268},
      archivePrefix={arXiv},
      primaryClass={cs.LG},
      url={https://arxiv.org/abs/2410.15268}, 
}
\bibliographystyle{icml2025}

%%%%%%%%%%%%%%%%%%%%%%%%%%%%%%%%%%%%%%%%%%%%%%%%%%%%%%%%%%%%%%%%%%%%%%%%%%%%%%%
%%%%%%%%%%%%%%%%%%%%%%%%%%%%%%%%%%%%%%%%%%%%%%%%%%%%%%%%%%%%%%%%%%%%%%%%%%%%%%%
% APPENDIX
%%%%%%%%%%%%%%%%%%%%%%%%%%%%%%%%%%%%%%%%%%%%%%%%%%%%%%%%%%%%%%%%%%%%%%%%%%%%%%%
%%%%%%%%%%%%%%%%%%%%%%%%%%%%%%%%%%%%%%%%%%%%%%%%%%%%%%%%%%%%%%%%%%%%%%%%%%%%%%%
\newpage
\appendix
\onecolumn

\section*{Appendix}

\section{Additional Related work}

\subsection{Concept Bottleneck Model}
\label{sec::related_cbm}
Concept Bottleneck Models (CBMs) aim to improve model transparency by first mapping inputs into an interpretable set of human-defined concepts (the concept bottleneck), and then making predictions based on those concepts. The original CBM framework~\cite{koh2020conceptbottleneckmodels} is trained on datasets where each input is annotated with both class labels and corresponding concept labels. At test time, the model predicts concepts from the input and uses them as intermediate representations to produce the final output via a classifier or regressor. This process enhances interpretability and enables human intervention by allowing concept-level edits. However, original CBM~\cite{koh2020conceptbottleneckmodels} requires substantial human effort to define the concept space and annotate each training sample with concept labels, which can be both time-consuming and labor-intensive. Moreover, they often suffer from suboptimal predictive performance. To address these limitations, Yuksekgonul et al.~\cite{yuksekgonul2023posthocCBM} propose a post-hoc CBM that converts any pretrained model into a concept bottleneck model. Their approach leverages multimodal approaches such as CLIP~\cite{radford2021learningtransferablevisualmodels} to align the input space (e.g., images) with a concept space (e.g., text), thereby reducing the need for explicitly labeled concept data. Nevertheless, this method still requires human expertise or additional learning steps to define the concept subspace. In a concurrent work, Oikarinen et al.~\cite{oikarinen2023labelfreeconceptbottleneckmodels} build upon similar ideas but go further by proposing a label-free CBM. They also utilize CLIP’s image and text encoders to map inputs to concepts, while fully automating the construction of the concept space using large language models (LLMs). Both approaches~\cite{oikarinen2023labelfreeconceptbottleneckmodels, yuksekgonul2023posthocCBM} report maintaining competitive predictive performance while improving interpretability.

In addition to these, several works explore specific challenges and extended settings of CBMs. Shang et al.~\cite{shang2024incrementalCBM} address the concept completeness problem by proposing to recover missing concepts through transforming complemented vectors with unclear semantics into potential concepts.  
Shin et al.~\cite{shin2023closerlookCBM} conduct in-depth analyses of intervention strategies in CBMs; for instance, they investigate which concept selection criteria are most cost-efficient yet effective in improving task performance.  
Kim et al.~\cite{kim2023probabilisticCBM} propose a probabilistic Concept Bottleneck Model to tackle ambiguity in concept prediction, which can undermine model reliability. Their approach explicitly models uncertainty in the concept space and provides explanations incorporating both the predicted concepts and their associated uncertainties.  
It is also worth mentioning that Xu et al.~\cite{xu2025graph} introduce a Graph Concept Bottleneck Model that facilitates the modeling of concept relationships by constructing a graph of latent concepts. Although it shares a similar name with our model, it tackles fundamentally different challenges.
All of the aforementioned works focus on Euclidean input spaces such as images, and how to adapt Concept Bottleneck Models to graph data remains largely unexplored.

\section{Additional details on methodology}
\label{addtional_detail}
\subsection{Prompt details}

\begin{tcolorbox}[colback=gray!10,colframe=gray!50!black,title=Prompt for self-supervised concept annotations]
\label{prompt:concept-annotation}
\texttt{Given \{graphML\} and \{dataset-details\}. \setlist{nolistsep}
\begin{enumerate} \itemsep0em 
\item Provide summary and context analysis on the graph. 
\item Identify a list of key concepts and themes presented in the graph. 
\end{enumerate}
}

\medskip
\texttt{GraphML} refers to the graph markup language~\cite{brandes2000graph} used for describing the graph (ego-net). We sample up to 20 neighboring nodes to control the prompt length.
\texttt{dataset-details} provides a detailed description of the graph dataset, including what each node/edge represents and relevant contextual information.

\end{tcolorbox}

\begin{tcolorbox}[colback=gray!10,colframe=gray!50!black,title=Prompt for Global Concept Proposal]
\texttt{In the domain of \{dataset-domain\}, list the related concepts/keywords for classifying the item as \{category\}.}

\medskip
\texttt{Dataset-domain} briefly describes the dataset’s domain or context, and \texttt{category} is the name of a class label from the downstream classification task. We apply this prompt to each class label and aggregate the generated concepts to form the initial concept pool.
\label{prompt:global-concept}
\end{tcolorbox}

\begin{tcolorbox}[colback=gray!10,colframe=gray!50!black,title=Prompt for Instance-Based Concept Extraction]
\texttt{Given a \{\texttt{graphML}\} and \{dataset-details\}. \setlist{nolistsep}
\begin{enumerate}[noitemsep]
\item Provide summary and context analysis on the graph. 
\item Identify a list of key concepts presented in the graph that are most important for determining its classification within the \{dataset-domain\}, which includes the following categories: \{category-list\}. 
\end{enumerate}
}

\medskip
\texttt{GraphML} refers to the graph markup language used for describing the graph (or ego-net if the instance is a node). \texttt{dataset-details} provides a detailed description of the graph dataset, \texttt{dataset-domain} briefly describes the dataset’s domain or context. \texttt{category-list} is the complete list of categories for the classification task to guide the LLM toward generating concepts that are helpful in predicting class labels. Only the outputted concept list from the second step is collected.
\label{prompt::local-concept}
\end{tcolorbox}

\subsection{Detailed procedures}
\label{append::instance-based}
\textbf{Instance-Based Concept Extraction}. We sample $m$ graph instances from each class and apply the prompt to each sampled graph instance, resulting in a large set of candidate concepts. We then identify a subset of concepts that are highly relevant to each class, distinct from those used by other classes, and useful for improving class discrimination. Specifically, for each class $y$, we calculate the class-wise concept activation score as:
\begin{equation}
    \bar{C}_y = \frac{1}{|\mathcal{D}_y|} \sum_{x_i \in \mathcal{D}_y} C_i,
\end{equation}
where $\mathcal{D}_y$ denotes the set of instances belonging to class $y$, and $C_i$ is the concept activation vector for instance $x_i$. Each element $C_i^{(j)}$ represents the activation score (e.g., cosine similarity) between the instance representation $f_{\theta}^{\text{GNN}}(x_i)$ and the embedding of the $j$-th concept $f^{\text{LM}}(c_j)$.

We then compute the discriminative score of concept $j$ for class $y$ as:
\begin{equation}
    \text{score}_j(y) = \bar{C}_y^{(j)} - \frac{1}{|\mathcal{Y}| - 1} \sum_{y' \ne y} \bar{C}_{y'}^{(j)},
\end{equation}
where $\mathcal{Y}$ is the set of all class labels, and $\bar{C}_y^{(j)}$ denotes the average activation of concept $j$ for class $y$. 

Finally, for each class, we select the top-$k$ concepts with the highest discriminative scores:
\begin{equation}
    \mathcal{C}^{\text{inst}} = \text{Top-$k$}_{j}(\text{score}_j(y)).
\end{equation}

\textbf{Details of Concept Filtering Process.}
\label{append::filtering}
Following similar procedures to \cite{oikarinen2023labelfreeconceptbottleneckmodels}, we apply a post-processing pipeline to refine the set of candidate concepts. The pipeline consists of the following steps:

\textit{(1) Removing overly long concepts.} Long concepts may reduce both interpretability and generalizability. We therefore tokenize each concept and discard those containing more than 10 tokens.

\textit{(2) Removing concepts overly similar to class labels.} Concepts that are identical or highly similar to class labels undermine the purpose of explanation. To mitigate this issue, we compute the cosine similarity between the Sentence-BERT embeddings of each concept and each class label, and filter out any concept with a similarity score greater than 0.85.

\textit{(3) Removing redundant concepts.} To reduce redundancy, we compute pairwise cosine similarity among concepts and remove any concept whose similarity with a retained concept exceeds 0.85.

\section{Supplemental experiment setups}\label{appendix:expt}
\subsection{Details of the datasets}
\label{sec::datasets}
In this section, we summarize the basic statistics of the datasets in our experimental evaluation in Table~\ref{tab:data}. All datasets used in our study are publicly available and come from diverse domains, including social media networks, citation graphs, and e-commerce graphs. Each node is associated with a class label, and most datasets contain more than two classes. The class distributions are imbalanced in these datasets. Therefore, in our experiments, we report node classification performance using the (Macro-)F1 score and balanced accuracy (BACC). For all datasets and settings, we adopt a default
train/validation/test split of 20\%/20\%/50\%, with the remain-
ing 10\% held out to investigate the effect of increased training data in later experiments. We use an inductive splitting,
where test nodes are entirely unseen during training.
\begin{table}
\centering
\scriptsize
% \fontsize{6.5pt}{7pt}\selectfont
\caption{Summary statistics of source and target datasets. }\label{tab:data}
% \vspace{-1em}
% \begin{tabular}{@{}l@{\hskip 4pt}r@{\hskip 4pt}r@{\hskip 4pt}l@{\hskip 4pt}l@{\hskip 4pt}r@{}}
\begin{tabular}{llllll}
\toprule
Dataset & \#Nodes & \#Edges & Type & Domain & \#Class \\
\midrule
Computers       & 87{,}229  & 721{,}081  & Co-purchase & E-commerce       & 10 \\
PubMed          & 19{,}717  & 44{,}338   & Citation     & Biomedicine      & 3  \\
Books-History   & 41{,}551  & 358{,}574  & Co-purchase & E-commerce       & 12 \\
Books-Children  & 76{,}875  & 1{,}554{,}578 & Co-purchase & E-commerce       & 24 \\
Sports-Fitness  & 173{,}055 & 1{,}773{,}500 & Co-purchase & E-commerce       & 13 \\
\midrule
Cora            & 2{,}708   & 5{,}429    & Citation     & Computer Science & 7  \\
CiteSeer        & 3{,}186   & 4{,}277    & Citation     & Computer Science & 6  \\
Instagram       & 11{,}339  & 144{,}010  & User-Post    & Social Media     & 2  \\
Reddit          & 33{,}434  & 198{,}448  & Post-Comment & Social Media     & 2  \\
WikiCS          & 11{,}701  & 215{,}863  & Article Link & Wikipedia        & 10 \\
\bottomrule
\end{tabular}
% \vspace{-2em}
\end{table}

\subsection{Implementation details}\label{sec::implementations} For all GNN-based methods, including those that use GNNs as backbones, we set the hidden dimension to 64 and the number of GNN layers to 2. For GAT and GT models, we use 4 attention heads. For the Set2set contrastive learning, we set the number of augmented views as 10, and temperature $\tau$ as 0.07. We ask the LLM to generate 10 concepts per instance and select top-100 concepts with highest relevance. We use the pretrained Sentence-BERT~\cite{reimers-2019-sentence-bert} model as the text encoder for all baselines.
For SEGNN~\cite{dai2021-SEGNN}, the original implementation requires access to all training nodes at test time in order to identify the closest neighbors and make predictions based on their labels. However, this approach is incompatible with our inductive setting, where the model is not permitted to access training instances during inference. To address this, we modify the implementation by introducing a small memory buffer that stores 5 randomly selected nodes per class from the training set. During testing, the model is restricted to retrieving neighbors only from this buffer.
For all self-explainable graph learning baselines, we follow the default hyperparameter settings provided in their open-source implementations. All experiments are conducted on four NVIDIA L40S GPUs. We access GPT-3.5 via the OpenAI API and set the temperature to 0 during graph summary and concept generation to avoid randomness.

\section{Complexity analysis}
\label{sec::complexity}

The primary overhead of \model{} lies in the pretraining stage, where a graph encoder is aligned with a semantically meaningful concept space using large language models (LLMs). However, this pretraining is performed once and can be reused across downstream datasets without incurring additional cost.
During the main training phase, where we optimize the information bottleneck criteria, the dominant cost comes from computing the gate vector $g$ (via a lightweight MLP) and training the classifier $\text{MLP}^{\text{cls}}$ on the masked concept representations. This results in a per-step complexity of $O(BKH)$, where $B$ is the batch size, $K$ is the number of candidate concepts, and $H$ is the hidden dimension of the MLP. The final predictor, after concept selection, operates on a reduced concept set and is simply a small MLP, which is highly efficient in both training and inference.
At inference time, \model{} consists of a frozen graph encoder (e.g., a GNN) followed by a fixed MLP classifier over selected concepts, making its runtime complexity comparable to that of a standard GNN model.

\section{Additional results}
\label{sec::addtional}
\subsection{Robustness Evaluation}

We report the results for all additional upsampling ratios $\gamma \in \{2, 3, 10\}$ in Table~\ref{tab:res_ood_2}, Table~\ref{tab:res_ood_3}, and Table~\ref{tab:res_ood_10}, respectively. Results for different perturbation ratios $\rho \in \{0.05, 0.1, 0.2, 0.5\}$ are shown in Table~\ref{tab:res_pertube_5}, Table~\ref{tab:res_pertube_10}, Table~\ref{tab:res_pertube_20}, and Table~\ref{tab:res_pertube_50}.

We emphasize that the test splits used for different upsampling ratios are not aligned, making direct comparison across these settings inappropriate. While a larger upsampling ratio increases the distribution shift between the training and test sets, it may also lead to a more balanced class distribution in the training or test data, which can sometimes improve test performance.
Regarding the perturbation setting, we observe that \model{} is the least affected by structural perturbation. We attribute this to the use of a fixed pretrained encoder, which is not updated during task-specific training. As a result, perturbing the training graph does not alter the graph embedding function. Moreover, the data augmentation used during pretraining also contributes to \model{}'s robustness under structural noise.
Interestingly, across all baseline methods, we do not observe a consistent trend correlating performance with increasing perturbation ratio. One possible explanation is that, for perturbation-sensitive models, even a small perturbation (e.g., $\rho=0.05$) significantly degrades performance, and the marginal impact of further perturbation is limited. Furthermore, recent studies such as \cite{han2023mlpinitembarrassinglysimplegnn} have shown that some GNNs can perform well even when trained without graph structure—effectively functioning like MLPs—and still generalize well when tested with full graph connectivity. When the perturbation ratio is large, models may similarly learn to disregard noisy structure, exhibiting behavior consistent with such MLP-based approaches and mitigating the negative effects of edge pertubation.
 
\begin{table}
\centering
\scriptsize
\caption{Node classification performance in \textit{regular settings}. The best-performing interpretable GNN on each dataset is \underline{underlined}, and the overall best-performing method is \textbf{bolded}.}
% \yue{GCB only achieves the best in 3 datasets; }
\label{tab:res_regular}
\begin{tabular}{l@{\hskip 5pt}c@{\hskip 3pt}c@{\hskip 7pt}@{\hskip 0pt}c@{\hskip 3pt}c@{\hskip 7pt}@{\hskip 0pt}c@{\hskip 3pt}c@{\hskip 7pt}@{\hskip 0pt}c@{\hskip 3pt}c@{\hskip 7pt}@{\hskip 0pt}c@{\hskip 3pt}c}
\toprule
 & \multicolumn{2}{c}{\texttt{Cora}} & \multicolumn{2}{c}{\texttt{Citeseer}} & \multicolumn{2}{c}{\texttt{Instagram}} & \multicolumn{2}{c}{\texttt{Reddit}} & \multicolumn{2}{c}{\texttt{WikiCS}} \\
 \cmidrule(lr){2-3} \cmidrule(lr){4-5} \cmidrule(lr){6-7} \cmidrule(lr){8-9} \cmidrule(lr){10-11} 
\textbf{Method} & F1 (\%) & BACC (\%) & F1 (\%) & BACC (\%) & F1 (\%) & BACC (\%) & F1 (\%) & BACC (\%) & F1 (\%) & BACC (\%) \\
\midrule
MLP & 68.00\textsubscript{(0.89)} & 67.49\textsubscript{(0.76)} & 63.81\textsubscript{(0.37)} & 64.29\textsubscript{(0.34)} & 53.78\textsubscript{(0.59)} & 53.76\textsubscript{(0.58)} & 53.34\textsubscript{(0.77)} & 53.39\textsubscript{(0.76)} & 69.22\textsubscript{(0.51)} & 69.25\textsubscript{(0.57)} \\
GCN & 72.38\textsubscript{(0.58)} & 71.95\textsubscript{(0.49)} & 62.47\textsubscript{(0.29)} & 63.05\textsubscript{(0.33)} & 53.63\textsubscript{(0.84)} & 53.62\textsubscript{(0.82)} & 54.49\textsubscript{(1.19)} & 54.64\textsubscript{(1.03)} & 67.45\textsubscript{(1.76)} & 68.84\textsubscript{(1.82)} \\
GAT & \textbf{74.58}\textsubscript{(0.95)} & \textbf{74.42}\textsubscript{(0.90)} & 63.92\textsubscript{(0.92)} & 64.47\textsubscript{(0.91)} & 55.71\textsubscript{(1.06)} & 55.74\textsubscript{(1.10)} & \textbf{56.29}\textsubscript{(0.60)} & 56.30\textsubscript{(0.59)} & 67.54\textsubscript{(1.68)} & 68.14\textsubscript{(1.59)} \\
SAGE & 70.59\textsubscript{(0.68)} & 70.66\textsubscript{(0.80)} & \textbf{65.09}\textsubscript{(0.62)} & \textbf{65.52}\textsubscript{(0.64)} & 54.49\textsubscript{(0.46)} & 54.50\textsubscript{(0.46)} & 55.33\textsubscript{(0.38)} & 55.33\textsubscript{(0.38)} & \textbf{72.96}\textsubscript{(0.30)} & \textbf{72.74}\textsubscript{(0.40)} \\
GT & 72.36\textsubscript{(1.96)} & 72.18\textsubscript{(1.56)} & 64.40\textsubscript{(0.76)} & 64.93\textsubscript{(0.68)} & 54.79\textsubscript{(0.40)} & 54.77\textsubscript{(0.39)} & 56.15\textsubscript{(0.39)} & 56.15\textsubscript{(0.39)} & 72.27\textsubscript{(0.52)} & 72.46\textsubscript{(0.56)} \\
\midrule
DIR-GNN & \underline{73.03}\textsubscript{(2.62)} & \underline{72.51}\textsubscript{(1.90)} & 62.10\textsubscript{(0.58)} & \underline{64.67}\textsubscript{(0.50)} & \underline{\textbf{56.76}}\textsubscript{(1.24)} & \underline{\textbf{57.37}}\textsubscript{(0.95)} & \underline{55.34}\textsubscript{(1.81)} & \underline{\textbf{57.18}}\textsubscript{(0.48)} & 67.14\textsubscript{(3.60)} & 66.26\textsubscript{(3.83)} \\
GIB & 66.81\textsubscript{(4.23)} & 67.23\textsubscript{(4.02)} & 49.28\textsubscript{(14.03)} & 53.88\textsubscript{(11.42)} & 40.72\textsubscript{(8.44)} & 51.52\textsubscript{(1.86)} & 38.84\textsubscript{(8.18)} & 51.49\textsubscript{(2.11)} & 45.30\textsubscript{(18.50)} & 45.38\textsubscript{(14.85)} \\
VGIB & 63.46\textsubscript{(28.19)} & 64.59\textsubscript{(25.11)} & 53.90\textsubscript{(19.24)} & 56.99\textsubscript{(16.88)} & 39.64\textsubscript{(1.64)} & 50.29\textsubscript{(0.58)} & 33.68\textsubscript{(1.13)} & 50.12\textsubscript{(0.22)} & 61.44\textsubscript{(25.27)} & 62.90\textsubscript{(22.46)} \\
SEGNN & 49.90\textsubscript{(4.09)} & 53.07\textsubscript{(3.30)} & 52.12\textsubscript{(5.51)} & 55.67\textsubscript{(4.11)} & 44.71\textsubscript{(2.56)} & 51.04\textsubscript{(0.49)} & 53.53\textsubscript{(1.66)} & 54.59\textsubscript{(0.90)} & 28.87\textsubscript{(3.57)} & 34.71\textsubscript{(2.78)} \\
    \model{} & 70.54\textsubscript{(1.33)} & 71.41\textsubscript{(0.88)} & \underline{63.22}\textsubscript{(0.50)} & 63.54\textsubscript{(0.49)} & \underline{\textbf{56.76}}\textsubscript{(0.55)} & 56.71\textsubscript{(0.51)} & 55.06\textsubscript{(0.72)} & 55.11\textsubscript{(0.72)} & \underline{68.82}\textsubscript{(0.41)} & \underline{70.64}\textsubscript{(0.82)} \\
\bottomrule
\end{tabular}
\end{table}

\begin{table}
    \centering
    \scriptsize
    \caption{Node classification performance in \textit{OOD settings} with upsampling ratio $\gamma=2$. The best-performing interpretable GNN on each dataset is \underline{underlined}, and the overall best-performing method is \textbf{bolded}.\newline}
    \begin{tabular}{l@{\hskip 5pt}c@{\hskip 3pt}c@{\hskip 7pt}@{\hskip 0pt}c@{\hskip 3pt}c@{\hskip 7pt}@{\hskip 0pt}c@{\hskip 3pt}c@{\hskip 7pt}@{\hskip 0pt}c@{\hskip 3pt}c@{\hskip 7pt}@{\hskip 0pt}c@{\hskip 3pt}c}
\toprule
 & \multicolumn{2}{c}{\textbf{Cora}} & \multicolumn{2}{c}{\textbf{Citeseer}} & \multicolumn{2}{c}{\textbf{Instagram}} & \multicolumn{2}{c}{\textbf{Reddit}} & \multicolumn{2}{c}{\textbf{WikiCS}} \\
 \cmidrule(lr){2-3} \cmidrule(lr){4-5} \cmidrule(lr){6-7} \cmidrule(lr){8-9} \cmidrule(lr){10-11} 
\textbf{Method} & F1 (\%) & BACC (\%) & F1 (\%) & BACC (\%) & F1 (\%) & BACC (\%) & F1 (\%) & BACC (\%) & F1 (\%) & BACC (\%) \\
\midrule
MLP & 46.52\textsubscript{(0.59)} & 57.50\textsubscript{(0.69)} & 44.89\textsubscript{(0.70)} & 60.49\textsubscript{(0.77)} & 35.55\textsubscript{(0.59)} & 51.54\textsubscript{(0.20)} & 17.03\textsubscript{(0.51)} & 51.28\textsubscript{(0.56)} & 53.82\textsubscript{(0.42)} & 63.23\textsubscript{(0.40)} \\
GCN & \textbf{56.10}\textsubscript{(0.46)} & \textbf{64.04}\textsubscript{(0.52)} & 41.06\textsubscript{(0.43)} & 56.08\textsubscript{(0.57)} & 39.36\textsubscript{(3.25)} & 52.54\textsubscript{(0.86)} & 16.65\textsubscript{(0.73)} & 50.74\textsubscript{(0.20)} & 53.80\textsubscript{(0.55)} & 60.37\textsubscript{(1.37)} \\
GAT & 52.54\textsubscript{(1.48)} & 63.29\textsubscript{(1.31)} & 44.71\textsubscript{(0.68)} & 60.63\textsubscript{(0.56)} & 33.42\textsubscript{(0.37)} & 51.49\textsubscript{(0.12)} & 13.06\textsubscript{(0.28)} & 49.78\textsubscript{(0.39)} & \textbf{56.41}\textsubscript{(2.24)} & 64.32\textsubscript{(1.67)} \\
SAGE & 40.39\textsubscript{(1.19)} & 50.69\textsubscript{(1.08)} & 40.99\textsubscript{(0.99)} & 56.02\textsubscript{(0.78)} & 35.71\textsubscript{(0.42)} & 51.76\textsubscript{(0.31)} & 15.97\textsubscript{(0.35)} & 50.74\textsubscript{(0.51)} & 49.65\textsubscript{(0.64)} & 60.20\textsubscript{(0.83)} \\
GT & 42.83\textsubscript{(1.35)} & 51.62\textsubscript{(1.59)} & 40.57\textsubscript{(1.22)} & 56.93\textsubscript{(0.90)} & 33.83\textsubscript{(0.60)} & 51.47\textsubscript{(0.28)} & 13.22\textsubscript{(0.33)} & 50.14\textsubscript{(0.34)} & 51.10\textsubscript{(0.69)} & 59.51\textsubscript{(0.99)} \\
DIR-GNN & 18.54\textsubscript{(2.90)} & 40.48\textsubscript{(2.24)} & 15.18\textsubscript{(0.70)} & 42.44\textsubscript{(0.52)} & 26.74\textsubscript{(0.00)} & 50.00\textsubscript{(0.00)} & 8.46\textsubscript{(0.00)} & 50.00\textsubscript{(0.00)} & 23.41\textsubscript{(1.92)} & 42.89\textsubscript{(0.55)} \\
GIB & 21.45\textsubscript{(3.55)} & 40.93\textsubscript{(1.77)} & 16.98\textsubscript{(4.91)} & 43.81\textsubscript{(3.24)} & 26.76\textsubscript{(0.02)} & 50.01\textsubscript{(0.01)} & 8.53\textsubscript{(0.06)} & 49.99\textsubscript{(0.06)} & 23.93\textsubscript{(1.12)} & 41.10\textsubscript{(1.63)} \\
VGIB & 45.60\textsubscript{(4.00)} & 58.19\textsubscript{(2.70)} & 15.61\textsubscript{(1.75)} & 44.07\textsubscript{(1.00)} & 26.74\textsubscript{(0.00)} & 50.00\textsubscript{(0.00)} & 8.46\textsubscript{(0.00)} & 50.00\textsubscript{(0.00)} & 54.31\textsubscript{(0.90)} & 63.54\textsubscript{(0.66)} \\
SEGNN & 40.04\textsubscript{(2.47)} & 51.44\textsubscript{(2.36)} & 25.59\textsubscript{(2.76)} & 45.69\textsubscript{(1.58)} & 26.74\textsubscript{(0.00)} & 50.00\textsubscript{(0.00)} & 8.46\textsubscript{(0.00)} & 50.00\textsubscript{(0.00)} & 37.26\textsubscript{(1.18)} & 49.80\textsubscript{(0.91)} \\
\model{} & \underline{54.23}\textsubscript{(0.00)} & \underline{62.97}\textsubscript{(1.15)} & \underline{\textbf{57.46}}\textsubscript{(0.85)} & \underline{\textbf{65.48}}\textsubscript{(0.65)} & \underline{\textbf{53.20}}\textsubscript{(0.81)} & \underline{\textbf{55.89}}\textsubscript{(0.82)} & \underline{\textbf{43.74}}\textsubscript{(0.77)} & \underline{\textbf{57.99}}\textsubscript{(0.71)} & \underline{55.19}\textsubscript{(0.72)} & \underline{\textbf{66.36}}\textsubscript{(0.62)} \\
\bottomrule
\end{tabular}
\label{tab:res_ood_2}

\end{table}

\begin{table}
    \centering
    \scriptsize
    \caption{Node classification performance in \textit{OOD settings} with upsampling ratio $\gamma=3$. The best-performing interpretable GNN on each dataset is \underline{underlined}, and the overall best-performing method is \textbf{bolded}.\newline}% (averaged over 5 trials). Standard deviation is denoted after $\pm$.}
\label{tab:res_ood_3}
\begin{tabular}{l@{\hskip 5pt}c@{\hskip 3pt}c@{\hskip 7pt}@{\hskip 0pt}c@{\hskip 3pt}c@{\hskip 7pt}@{\hskip 0pt}c@{\hskip 3pt}c@{\hskip 7pt}@{\hskip 0pt}c@{\hskip 3pt}c@{\hskip 7pt}@{\hskip 0pt}c@{\hskip 3pt}c}
\toprule
 & \multicolumn{2}{c}{\textbf{Cora}} & \multicolumn{2}{c}{\textbf{Citeseer}} & \multicolumn{2}{c}{\textbf{Instagram}} & \multicolumn{2}{c}{\textbf{Reddit}} & \multicolumn{2}{c}{\textbf{WikiCS}} \\
 \cmidrule(lr){2-3} \cmidrule(lr){4-5} \cmidrule(lr){6-7} \cmidrule(lr){8-9} \cmidrule(lr){10-11} 
\textbf{Method} & F1 (\%) & BACC (\%) & F1 (\%) & BACC (\%) & F1 (\%) & BACC (\%) & F1 (\%) & BACC (\%) & F1 (\%) & BACC (\%) \\
\midrule
MLP & 43.78\textsubscript{(0.52)} & 54.85\textsubscript{(0.57)} & 45.73\textsubscript{(0.59)} & 58.72\textsubscript{(0.71)} & 37.04\textsubscript{(0.67)} & 52.33\textsubscript{(0.25)} & 16.37\textsubscript{(0.61)} & 51.37\textsubscript{(0.32)} & 55.01\textsubscript{(0.33)} & 65.33\textsubscript{(1.15)} \\
GCN & \textbf{57.28}\textsubscript{(1.35)} & \textbf{67.86}\textsubscript{(0.98)} & 41.59\textsubscript{(0.68)} & 56.90\textsubscript{(0.80)} & 37.73\textsubscript{(3.06)} & 51.58\textsubscript{(0.59)} & 16.00\textsubscript{(0.35)} & 49.23\textsubscript{(0.48)} & 54.99\textsubscript{(1.28)} & 61.65\textsubscript{(2.33)} \\
GAT & 52.81\textsubscript{(1.43)} & 60.66\textsubscript{(1.28)} & 43.09\textsubscript{(1.45)} & 58.16\textsubscript{(1.39)} & 34.77\textsubscript{(1.08)} & 51.50\textsubscript{(0.38)} & 13.85\textsubscript{(0.44)} & 49.44\textsubscript{(0.28)} & \textbf{56.99}\textsubscript{(0.10)} & 65.85\textsubscript{(0.84)} \\
SAGE & 51.40\textsubscript{(1.77)} & 62.46\textsubscript{(1.54)} & 36.25\textsubscript{(1.39)} & 52.63\textsubscript{(1.14)} & 35.04\textsubscript{(0.55)} & 51.47\textsubscript{(0.27)} & 14.36\textsubscript{(0.17)} & 48.96\textsubscript{(0.40)} & 52.43\textsubscript{(0.78)} & 60.92\textsubscript{(0.97)} \\
GT & 47.70\textsubscript{(1.31)} & 58.26\textsubscript{(1.19)} & 36.12\textsubscript{(1.62)} & 53.10\textsubscript{(1.23)} & 33.16\textsubscript{(0.47)} & 50.15\textsubscript{(0.37)} & 14.18\textsubscript{(0.15)} & 49.59\textsubscript{(0.21)} & 54.83\textsubscript{(0.89)} & 62.58\textsubscript{(1.07)} \\
DIR-GNN & 20.13\textsubscript{(2.75)} & 41.89\textsubscript{(1.50)} & 14.70\textsubscript{(0.34)} & 43.15\textsubscript{(0.38)} & 26.74\textsubscript{(0.00)} & 50.00\textsubscript{(0.00)} & 8.46\textsubscript{(0.00)} & 50.00\textsubscript{(0.00)} & 23.60\textsubscript{(1.40)} & 42.84\textsubscript{(0.57)} \\
GIB & 22.43\textsubscript{(5.91)} & 41.53\textsubscript{(4.24)} & 17.72\textsubscript{(5.88)} & 44.14\textsubscript{(4.31)} & 26.74\textsubscript{(0.00)} & 50.00\textsubscript{(0.01)} & 8.48\textsubscript{(0.04)} & 50.01\textsubscript{(0.02)} & 20.30\textsubscript{(7.70)} & 35.16\textsubscript{(9.59)} \\
VGIB & 44.05\textsubscript{(2.53)} & 57.14\textsubscript{(1.92)} & 17.14\textsubscript{(4.35)} & 44.50\textsubscript{(2.44)} & 26.74\textsubscript{(0.00)} & 50.00\textsubscript{(0.00)} & 8.46\textsubscript{(0.00)} & 50.00\textsubscript{(0.00)} & 54.74\textsubscript{(1.32)} & 63.15\textsubscript{(1.20)} \\
SEGNN & 29.92\textsubscript{(1.05)} & 46.62\textsubscript{(0.81)} & 25.15\textsubscript{(9.17)} & 43.70\textsubscript{(6.98)} & 26.74\textsubscript{(0.00)} & 50.00\textsubscript{(0.00)} & 8.46\textsubscript{(0.00)} & 50.00\textsubscript{(0.00)} & 27.85\textsubscript{(1.02)} & 43.33\textsubscript{(1.50)} \\
\model{} & \underline{54.99}\textsubscript{(0.00)} & \underline{65.00}\textsubscript{(0.00)} & \underline{\textbf{57.85}}\textsubscript{(0.27)} & \underline{\textbf{65.62}}\textsubscript{(0.79)} & \underline{\textbf{54.54}}\textsubscript{(0.17)} & \underline{\textbf{56.39}}\textsubscript{(0.36)} & \underline{\textbf{45.82}}\textsubscript{(0.31)} & \underline{\textbf{60.65}}\textsubscript{(0.83)} & \underline{54.97}\textsubscript{(0.37)} & \underline{\textbf{66.70}}\textsubscript{(0.40)} \\
\bottomrule
\end{tabular}
\end{table}

\begin{table}
    \centering
    \scriptsize
    \caption{Node classification performance in \textit{OOD settings} with upsampling ratio $\gamma=10$. The best-performing interpretable GNN on each dataset is \underline{underlined}, and the overall best-performing method is \textbf{bolded}.\newline}% (averaged over 5 trials). Standard deviation is denoted after $\pm$.}
\label{tab:res_ood_10}
\begin{tabular}{l@{\hskip 5pt}c@{\hskip 3pt}c@{\hskip 7pt}@{\hskip 0pt}c@{\hskip 3pt}c@{\hskip 7pt}@{\hskip 0pt}c@{\hskip 3pt}c@{\hskip 7pt}@{\hskip 0pt}c@{\hskip 3pt}c@{\hskip 7pt}@{\hskip 0pt}c@{\hskip 3pt}c}
\toprule
 & \multicolumn{2}{c}{\textbf{Cora}} & \multicolumn{2}{c}{\textbf{Citeseer}} & \multicolumn{2}{c}{\textbf{Instagram}} & \multicolumn{2}{c}{\textbf{Reddit}} & \multicolumn{2}{c}{\textbf{WikiCS}} \\
 \cmidrule(lr){2-3} \cmidrule(lr){4-5} \cmidrule(lr){6-7} \cmidrule(lr){8-9} \cmidrule(lr){10-11} 
\textbf{Method} & F1 (\%) & BACC (\%) & F1 (\%) & BACC (\%) & F1 (\%) & BACC (\%) & F1 (\%) & BACC (\%) & F1 (\%) & BACC (\%) \\
\midrule
MLP & 47.58\textsubscript{(0.44)} & 59.14\textsubscript{(0.61)} & 41.44\textsubscript{(0.42)} & 56.87\textsubscript{(0.70)} & 35.38\textsubscript{(0.66)} & 51.29\textsubscript{(0.43)} & 16.08\textsubscript{(0.42)} & 50.69\textsubscript{(0.34)} & 52.72\textsubscript{(0.50)} & 62.79\textsubscript{(0.50)} \\
GCN & \textbf{62.08}\textsubscript{(1.59)} & 69.26\textsubscript{(1.83)} & 43.58\textsubscript{(0.45)} & 54.79\textsubscript{(0.34)} & 47.15\textsubscript{(4.10)} & 55.23\textsubscript{(1.48)} & 17.35\textsubscript{(0.97)} & 49.91\textsubscript{(0.17)} & 62.47\textsubscript{(1.15)} & 64.89\textsubscript{(1.38)} \\
GAT & 60.32\textsubscript{(1.56)} & 68.94\textsubscript{(1.15)} & 48.46\textsubscript{(0.66)} & 61.74\textsubscript{(0.80)} & 35.80\textsubscript{(0.67)} & 51.70\textsubscript{(0.30)} & 15.35\textsubscript{(1.35)} & 51.40\textsubscript{(0.38)} & 57.17\textsubscript{(1.59)} & 60.82\textsubscript{(1.69)} \\
SAGE & 50.49\textsubscript{(1.06)} & 57.96\textsubscript{(1.12)} & 35.75\textsubscript{(1.49)} & 53.22\textsubscript{(0.90)} & 37.94\textsubscript{(0.94)} & 52.14\textsubscript{(0.40)} & 16.70\textsubscript{(0.73)} & 52.08\textsubscript{(0.22)} & \textbf{62.88}\textsubscript{(0.18)} & \textbf{72.40}\textsubscript{(1.02)} \\
GT & 47.31\textsubscript{(2.61)} & 56.27\textsubscript{(1.92)} & 30.80\textsubscript{(1.22)} & 50.68\textsubscript{(0.82)} & 37.51\textsubscript{(0.46)} & 52.90\textsubscript{(0.16)} & 17.33\textsubscript{(0.79)} & 51.58\textsubscript{(0.24)} & 62.18\textsubscript{(0.80)} & 68.79\textsubscript{(1.87)} \\
DIR-GNN & 22.04\textsubscript{(3.39)} & 42.60\textsubscript{(2.67)} & 15.56\textsubscript{(1.05)} & 42.93\textsubscript{(0.37)} & 26.74\textsubscript{(0.00)} & 50.00\textsubscript{(0.00)} & 8.46\textsubscript{(0.00)} & 50.00\textsubscript{(0.00)} & 23.89\textsubscript{(0.58)} & 42.04\textsubscript{(0.65)} \\
GIB & 26.30\textsubscript{(8.89)} & 44.06\textsubscript{(5.96)} & 14.94\textsubscript{(0.54)} & 42.42\textsubscript{(0.68)} & 26.92\textsubscript{(0.31)} & 50.06\textsubscript{(0.13)} & 8.46\textsubscript{(0.02)} & 49.98\textsubscript{(0.05)} & 25.39\textsubscript{(2.25)} & 38.19\textsubscript{(1.06)} \\
VGIB & 60.87\textsubscript{(3.20)} & 69.42\textsubscript{(3.04)} & 24.29\textsubscript{(6.80)} & 48.20\textsubscript{(3.96)} & 26.74\textsubscript{(0.00)} & 50.00\textsubscript{(0.00)} & 8.46\textsubscript{(0.00)} & 50.00\textsubscript{(0.00)} & 61.85\textsubscript{(1.81)} & 69.02\textsubscript{(1.35)} \\
\model{} & \underline{61.68}\textsubscript{(1.63)} & \underline{\textbf{69.55}}\textsubscript{(1.11)} & \underline{\textbf{58.08}}\textsubscript{(0.34)} & \underline{\textbf{65.52}}\textsubscript{(0.25)} & \underline{\textbf{52.17}}\textsubscript{(2.34)} & \underline{\textbf{55.57}}\textsubscript{(1.15)} & \underline{\textbf{44.77}}\textsubscript{(1.34)} & \underline{\textbf{55.75}}\textsubscript{(0.43)} & \underline{60.66}\textsubscript{(0.97)} & \underline{71.22}\textsubscript{(1.43)} \\
\bottomrule
\end{tabular}
\end{table}

\begin{table}
    \centering
    \scriptsize
        \caption{Node classification performance in \textit{adversarial settings} with perturbation ratio $\rho=0.05$. The best-performing interpretable GNN on each dataset is \underline{underlined}, and the overall best-performing method is \textbf{bolded}.\newline}% (averaged over 5 trials). Standard deviation is denoted after $\pm$.}
\label{tab:res_pertube_5}
\begin{tabular}{l@{\hskip 5pt}c@{\hskip 3pt}c@{\hskip 7pt}@{\hskip 0pt}c@{\hskip 3pt}c@{\hskip 7pt}@{\hskip 0pt}c@{\hskip 3pt}c@{\hskip 7pt}@{\hskip 0pt}c@{\hskip 3pt}c@{\hskip 7pt}@{\hskip 0pt}c@{\hskip 3pt}c}
\toprule
 & \multicolumn{2}{c}{\textbf{Cora}} & \multicolumn{2}{c}{\textbf{Citeseer}} & \multicolumn{2}{c}{\textbf{Instagram}} & \multicolumn{2}{c}{\textbf{Reddit}} & \multicolumn{2}{c}{\textbf{WikiCS}} \\
 \cmidrule(lr){2-3} \cmidrule(lr){4-5} \cmidrule(lr){6-7} \cmidrule(lr){8-9} \cmidrule(lr){10-11} 
\textbf{Method} & F1 (\%) & BACC (\%) & F1 (\%) & BACC (\%) & F1 (\%) & BACC (\%) & F1 (\%) & BACC (\%) & F1 (\%) & BACC (\%) \\
\midrule
MLP & 46.79\textsubscript{(0.88)} & 58.83\textsubscript{(0.84)} & 38.16\textsubscript{(0.41)} & 53.71\textsubscript{(0.37)} & 37.69\textsubscript{(0.44)} & 52.53\textsubscript{(0.15)} & 16.19\textsubscript{(0.46)} & 51.58\textsubscript{(0.44)} & 53.96\textsubscript{(0.34)} & 64.83\textsubscript{(0.25)} \\
GCN & 58.93\textsubscript{(1.32)} & 67.16\textsubscript{(1.38)} & 46.96\textsubscript{(0.95)} & 58.48\textsubscript{(1.13)} & 42.47\textsubscript{(4.66)} & 52.11\textsubscript{(1.09)} & 15.95\textsubscript{(0.86)} & 50.46\textsubscript{(0.68)} & 62.44\textsubscript{(0.37)} & 68.39\textsubscript{(0.49)} \\
GAT & 55.01\textsubscript{(1.93)} & 61.57\textsubscript{(1.94)} & 43.59\textsubscript{(1.57)} & 57.78\textsubscript{(1.21)} & 34.32\textsubscript{(1.37)} & 51.16\textsubscript{(0.53)} & 17.03\textsubscript{(0.95)} & 51.28\textsubscript{(0.42)} & 58.93\textsubscript{(2.81)} & 66.43\textsubscript{(2.35)} \\
SAGE & 53.45\textsubscript{(2.23)} & 57.21\textsubscript{(2.34)} & 42.45\textsubscript{(1.44)} & 57.74\textsubscript{(0.95)} & 40.93\textsubscript{(6.08)} & 52.32\textsubscript{(0.67)} & 16.00\textsubscript{(0.61)} & 51.42\textsubscript{(0.61)} & 61.58\textsubscript{(0.22)} & 68.96\textsubscript{(1.30)} \\
GT & 38.25\textsubscript{(2.04)} & 45.19\textsubscript{(1.18)} & 39.80\textsubscript{(3.32)} & 55.30\textsubscript{(2.16)} & 34.43\textsubscript{(1.15)} & 51.44\textsubscript{(0.30)} & 14.35\textsubscript{(0.74)} & 51.02\textsubscript{(0.27)} & 56.30\textsubscript{(1.16)} & 64.35\textsubscript{(1.49)} \\
DIR-GNN & \underline{\textbf{73.48}}\textsubscript{(1.08)} & \underline{\textbf{72.72}}\textsubscript{(1.37)} & 62.03\textsubscript{(0.64)} & \underline{\textbf{64.60}}\textsubscript{(0.54)} & 55.78\textsubscript{(2.54)} & 56.70\textsubscript{(1.42)} & 54.64\textsubscript{(2.70)} & 57.12\textsubscript{(1.00)} & 65.05\textsubscript{(1.45)} & 63.77\textsubscript{(1.50)} \\
GIB & 58.60\textsubscript{(15.18)} & 59.17\textsubscript{(14.38)} & 45.60\textsubscript{(17.26)} & 50.91\textsubscript{(13.49)} & 40.96\textsubscript{(8.68)} & 51.59\textsubscript{(1.93)} & 38.57\textsubscript{(7.79)} & 51.70\textsubscript{(2.56)} & 40.07\textsubscript{(16.67)} & 40.14\textsubscript{(12.96)} \\
VGIB & 21.17\textsubscript{(26.63)} & 26.65\textsubscript{(23.68)} & 53.90\textsubscript{(19.09)} & 57.17\textsubscript{(16.80)} & 38.99\textsubscript{(0.34)} & 50.07\textsubscript{(0.14)} & 34.58\textsubscript{(2.56)} & 50.29\textsubscript{(0.47)} & \underline{\textbf{72.78}}\textsubscript{(1.07)} & \underline{\textbf{72.45}}\textsubscript{(1.37)} \\
SEGNN & 55.79\textsubscript{(1.48)} & 59.39\textsubscript{(1.03)} & 60.06\textsubscript{(0.69)} & 62.95\textsubscript{(0.74)} & 54.75\textsubscript{(1.01)} & 55.22\textsubscript{(0.91)} & 55.58\textsubscript{(0.36)} & \underline{\textbf{55.99}}\textsubscript{(0.30)} & 37.35\textsubscript{(0.71)} & 41.85\textsubscript{(1.12)} \\
\model{} & 70.75\textsubscript{(0.85)} & 71.34\textsubscript{(1.05)} & \underline{\textbf{63.20}}\textsubscript{(0.76)} & 63.52\textsubscript{(0.78)} & \underline{\textbf{56.79}}\textsubscript{(0.60)} & \underline{\textbf{56.72}}\textsubscript{(0.59)} & \underline{\textbf{54.93}}\textsubscript{(0.78)} & 54.98\textsubscript{(0.79)} & 68.70\textsubscript{(0.42)} & 70.60\textsubscript{(0.46)} \\
\bottomrule
\end{tabular}
\end{table}

\begin{table}
    \centering
    \scriptsize
        \caption{Node classification performance in \textit{adversarial settings} with perturbation ratio $\rho=0.1$. The best-performing interpretable GNN on each dataset is \underline{underlined}, and the overall best-performing method is \textbf{bolded}.\newline}% (averaged over 5 trials). Standard deviation is denoted after $\pm$.}
\label{tab:res_pertube_10}
\begin{tabular}{l@{\hskip 5pt}c@{\hskip 3pt}c@{\hskip 7pt}@{\hskip 0pt}c@{\hskip 3pt}c@{\hskip 7pt}@{\hskip 0pt}c@{\hskip 3pt}c@{\hskip 7pt}@{\hskip 0pt}c@{\hskip 3pt}c@{\hskip 7pt}@{\hskip 0pt}c@{\hskip 3pt}c}
\toprule
 & \multicolumn{2}{c}{\textbf{Cora}} & \multicolumn{2}{c}{\textbf{Citeseer}} & \multicolumn{2}{c}{\textbf{Instagram}} & \multicolumn{2}{c}{\textbf{Reddit}} & \multicolumn{2}{c}{\textbf{WikiCS}} \\
 \cmidrule(lr){2-3} \cmidrule(lr){4-5} \cmidrule(lr){6-7} \cmidrule(lr){8-9} \cmidrule(lr){10-11} 
\textbf{Method} & F1 (\%) & BACC (\%) & F1 (\%) & BACC (\%) & F1 (\%) & BACC (\%) & F1 (\%) & BACC (\%) & F1 (\%) & BACC (\%) \\
\midrule
MLP & 45.04\textsubscript{(1.20)} & 57.94\textsubscript{(0.91)} & 41.94\textsubscript{(0.33)} & 57.38\textsubscript{(0.21)} & 34.65\textsubscript{(0.60)} & 51.57\textsubscript{(0.30)} & 18.09\textsubscript{(0.43)} & 51.09\textsubscript{(0.58)} & 54.13\textsubscript{(0.30)} & 64.72\textsubscript{(0.64)} \\
GCN & 65.19\textsubscript{(1.66)} & 67.62\textsubscript{(1.61)} & 46.43\textsubscript{(1.13)} & 58.40\textsubscript{(1.17)} & 38.56\textsubscript{(1.38)} & 51.96\textsubscript{(0.60)} & 18.17\textsubscript{(1.20)} & 50.51\textsubscript{(0.67)} & 63.15\textsubscript{(2.44)} & 68.98\textsubscript{(2.14)} \\
GAT & 63.56\textsubscript{(1.59)} & 68.88\textsubscript{(1.40)} & 43.79\textsubscript{(1.41)} & 58.11\textsubscript{(0.90)} & 35.94\textsubscript{(2.11)} & 51.83\textsubscript{(0.55)} & 17.30\textsubscript{(1.53)} & 51.83\textsubscript{(0.72)} & 58.79\textsubscript{(1.66)} & 67.83\textsubscript{(1.26)} \\
SAGE & 47.78\textsubscript{(0.92)} & 55.17\textsubscript{(0.52)} & 40.60\textsubscript{(0.80)} & 56.65\textsubscript{(0.63)} & 38.86\textsubscript{(1.28)} & 52.61\textsubscript{(0.62)} & 16.93\textsubscript{(0.48)} & 51.67\textsubscript{(0.34)} & 59.90\textsubscript{(0.67)} & 66.69\textsubscript{(0.61)} \\
GT & 40.32\textsubscript{(1.93)} & 46.77\textsubscript{(1.27)} & 27.14\textsubscript{(1.44)} & 46.19\textsubscript{(1.04)} & 35.26\textsubscript{(0.73)} & 51.95\textsubscript{(0.35)} & 16.80\textsubscript{(1.02)} & 51.26\textsubscript{(0.49)} & 60.67\textsubscript{(0.99)} & 67.86\textsubscript{(1.05)} \\
DIR-GNN & \underline{\textbf{71.70}}\textsubscript{(2.79)} & 71.04\textsubscript{(2.08)} & 61.84\textsubscript{(1.36)} & \underline{\textbf{64.42}}\textsubscript{(1.24)} & 55.55\textsubscript{(2.17)} & 56.66\textsubscript{(1.28)} & \underline{\textbf{55.41}}\textsubscript{(1.29)} & \underline{\textbf{57.48}}\textsubscript{(0.53)} & 64.30\textsubscript{(4.20)} & 63.15\textsubscript{(4.09)} \\
GIB & 55.55\textsubscript{(16.26)} & 58.38\textsubscript{(11.63)} & 58.99\textsubscript{(4.11)} & 61.91\textsubscript{(3.36)} & 41.53\textsubscript{(9.29)} & 51.81\textsubscript{(2.20)} & 40.23\textsubscript{(8.43)} & 52.11\textsubscript{(2.82)} & 30.36\textsubscript{(14.24)} & 32.54\textsubscript{(12.21)} \\
VGIB & 22.42\textsubscript{(26.34)} & 28.23\textsubscript{(23.26)} & 52.91\textsubscript{(22.73)} & 55.81\textsubscript{(19.19)} & 38.86\textsubscript{(0.07)} & 50.02\textsubscript{(0.03)} & 33.92\textsubscript{(1.58)} & 50.16\textsubscript{(0.30)} & 59.83\textsubscript{(24.76)} & 60.32\textsubscript{(22.57)} \\
SEGNN & 56.89\textsubscript{(0.75)} & 60.23\textsubscript{(0.64)} & 59.55\textsubscript{(0.62)} & 62.66\textsubscript{(0.62)} & 54.67\textsubscript{(0.70)} & 55.42\textsubscript{(0.77)} & 55.91\textsubscript{(1.85)} & 56.70\textsubscript{(1.09)} & 36.78\textsubscript{(1.67)} & 41.06\textsubscript{(1.82)} \\
\model{} & 70.54\textsubscript{(1.54)} & \underline{\textbf{71.31}}\textsubscript{(2.31)} & \underline{\textbf{63.02}}\textsubscript{(0.40)} & 63.38\textsubscript{(0.44)} & \underline{\textbf{56.75}}\textsubscript{(0.36)} & \underline{\textbf{56.70}}\textsubscript{(0.38)} & 54.91\textsubscript{(0.40)} & 54.95\textsubscript{(0.38)} & \underline{\textbf{68.80}}\textsubscript{(0.30)} & \underline{\textbf{70.45}}\textsubscript{(0.43)} \\
\bottomrule
\end{tabular}

\end{table}

\begin{table}
    \centering
    \scriptsize
        \caption{Node classification performance in \textit{adversarial settings} with perturbation ratio $\rho=0.2$. The best-performing interpretable GNN on each dataset is \underline{underlined}, and the overall best-performing method is \textbf{bolded}.\newline}% (averaged over 5 trials). Standard deviation is denoted after $\pm$.}
\label{tab:res_pertube_20}
\begin{tabular}{l@{\hskip 5pt}c@{\hskip 3pt}c@{\hskip 7pt}@{\hskip 0pt}c@{\hskip 3pt}c@{\hskip 7pt}@{\hskip 0pt}c@{\hskip 3pt}c@{\hskip 7pt}@{\hskip 0pt}c@{\hskip 3pt}c@{\hskip 7pt}@{\hskip 0pt}c@{\hskip 3pt}c}
\toprule
 & \multicolumn{2}{c}{\textbf{Cora}} & \multicolumn{2}{c}{\textbf{Citeseer}} & \multicolumn{2}{c}{\textbf{Instagram}} & \multicolumn{2}{c}{\textbf{Reddit}} & \multicolumn{2}{c}{\textbf{WikiCS}} \\
 \cmidrule(lr){2-3} \cmidrule(lr){4-5} \cmidrule(lr){6-7} \cmidrule(lr){8-9} \cmidrule(lr){10-11} 
\textbf{Method} & F1 (\%) & BACC (\%) & F1 (\%) & BACC (\%) & F1 (\%) & BACC (\%) & F1 (\%) & BACC (\%) & F1 (\%) & BACC (\%) \\
\midrule
MLP & 49.30\textsubscript{(0.81)} & 59.94\textsubscript{(0.97)} & 42.01\textsubscript{(0.43)} & 57.97\textsubscript{(0.22)} & 37.57\textsubscript{(3.23)} & 51.27\textsubscript{(0.41)} & 15.37\textsubscript{(0.09)} & 51.51\textsubscript{(0.29)} & 52.92\textsubscript{(0.41)} & 61.84\textsubscript{(0.70)} \\
GCN & 60.24\textsubscript{(0.83)} & 68.81\textsubscript{(1.01)} & 47.16\textsubscript{(1.29)} & 58.67\textsubscript{(1.02)} & 37.70\textsubscript{(2.16)} & 51.52\textsubscript{(0.53)} & 17.24\textsubscript{(1.82)} & 50.60\textsubscript{(0.84)} & 59.73\textsubscript{(0.71)} & 62.88\textsubscript{(0.35)} \\
GAT & 57.88\textsubscript{(2.24)} & 64.39\textsubscript{(1.85)} & 44.69\textsubscript{(1.35)} & 58.83\textsubscript{(1.13)} & 37.82\textsubscript{(1.08)} & 52.08\textsubscript{(0.46)} & 14.93\textsubscript{(1.04)} & 50.50\textsubscript{(0.56)} & 58.14\textsubscript{(2.12)} & 62.40\textsubscript{(2.18)} \\
SAGE & 50.26\textsubscript{(1.95)} & 57.82\textsubscript{(1.43)} & 29.81\textsubscript{(2.27)} & 49.99\textsubscript{(1.55)} & 36.98\textsubscript{(0.47)} & 52.00\textsubscript{(0.08)} & 17.10\textsubscript{(0.40)} & 50.99\textsubscript{(0.43)} & 62.87\textsubscript{(0.63)} & 70.36\textsubscript{(0.31)} \\
GT & 51.12\textsubscript{(2.34)} & 56.14\textsubscript{(2.39)} & 32.63\textsubscript{(1.41)} & 51.08\textsubscript{(0.71)} & 35.34\textsubscript{(0.86)} & 51.61\textsubscript{(0.50)} & 16.19\textsubscript{(0.68)} & 50.84\textsubscript{(0.28)} & 60.47\textsubscript{(0.47)} & 66.23\textsubscript{(0.77)} \\
DIR-GNN & \underline{\textbf{71.30}}\textsubscript{(2.36)} & \underline{\textbf{71.11}}\textsubscript{(1.94)} & 62.54\textsubscript{(0.34)} & \underline{\textbf{65.12}}\textsubscript{(0.38)} & 55.77\textsubscript{(2.23)} & \underline{\textbf{56.87}}\textsubscript{(1.32)} & 54.68\textsubscript{(2.36)} & \underline{\textbf{56.99}}\textsubscript{(0.90)} & 61.70\textsubscript{(3.35)} & 60.60\textsubscript{(3.45)} \\
GIB & 37.52\textsubscript{(19.90)} & 42.46\textsubscript{(16.75)} & 52.91\textsubscript{(12.84)} & 57.50\textsubscript{(9.10)} & 40.83\textsubscript{(8.60)} & 51.53\textsubscript{(1.89)} & 41.45\textsubscript{(9.60)} & 51.65\textsubscript{(2.13)} & 24.40\textsubscript{(11.79)} & 27.94\textsubscript{(10.41)} \\
VGIB & 34.11\textsubscript{(32.96)} & 38.47\textsubscript{(29.40)} & 52.05\textsubscript{(22.62)} & 55.80\textsubscript{(19.34)} & 40.36\textsubscript{(3.07)} & 50.41\textsubscript{(0.81)} & 33.09\textsubscript{(0.08)} & 50.00\textsubscript{(0.02)} & 59.54\textsubscript{(24.67)} & 61.20\textsubscript{(21.58)} \\
SEGNN & 55.76\textsubscript{(1.87)} & 59.44\textsubscript{(1.21)} & 59.94\textsubscript{(0.64)} & 62.98\textsubscript{(0.54)} & 55.07\textsubscript{(1.63)} & 55.27\textsubscript{(1.65)} & 54.56\textsubscript{(0.07)} & 55.35\textsubscript{(0.37)} & 35.77\textsubscript{(0.70)} & 40.40\textsubscript{(0.89)} \\
\model{} & 70.40\textsubscript{(1.32)} & 71.03\textsubscript{(0.60)} & \underline{\textbf{63.14}}\textsubscript{(0.74)} & 63.47\textsubscript{(0.70)} & \underline{\textbf{56.81}}\textsubscript{(0.28)} & 56.76\textsubscript{(0.30)} & \underline{\textbf{55.05}}\textsubscript{(0.44)} & 55.10\textsubscript{(0.45)} & \underline{\textbf{68.71}}\textsubscript{(0.55)} & \underline{\textbf{70.64}}\textsubscript{(0.59)} \\
\bottomrule
\end{tabular}

\end{table}

\begin{table}
    \centering
    \scriptsize
        \caption{Node classification performance in \textit{adversarial settings} with perturbation ratio $\rho=0.5$. The best-performing interpretable GNN on each dataset is \underline{underlined}, and the overall best-performing method is \textbf{bolded}.\newline}% (averaged over 5 trials). Standard deviation is denoted after $\pm$.}
\label{tab:res_pertube_50}
\begin{tabular}{l@{\hskip 5pt}c@{\hskip 3pt}c@{\hskip 7pt}@{\hskip 0pt}c@{\hskip 3pt}c@{\hskip 7pt}@{\hskip 0pt}c@{\hskip 3pt}c@{\hskip 7pt}@{\hskip 0pt}c@{\hskip 3pt}c@{\hskip 7pt}@{\hskip 0pt}c@{\hskip 3pt}c}
\toprule
 & \multicolumn{2}{c}{\textbf{Cora}} & \multicolumn{2}{c}{\textbf{Citeseer}} & \multicolumn{2}{c}{\textbf{Instagram}} & \multicolumn{2}{c}{\textbf{Reddit}} & \multicolumn{2}{c}{\textbf{WikiCS}} \\
 \cmidrule(lr){2-3} \cmidrule(lr){4-5} \cmidrule(lr){6-7} \cmidrule(lr){8-9} \cmidrule(lr){10-11} 
\textbf{Method} & F1 (\%) & BACC (\%) & F1 (\%) & BACC (\%) & F1 (\%) & BACC (\%) & F1 (\%) & BACC (\%) & F1 (\%) & BACC (\%) \\
\midrule
MLP & 47.76\textsubscript{(0.43)} & 58.52\textsubscript{(0.43)} & 44.65\textsubscript{(0.39)} & 58.45\textsubscript{(0.41)} & 35.26\textsubscript{(0.57)} & 51.72\textsubscript{(0.39)} & 16.78\textsubscript{(0.31)} & 50.46\textsubscript{(0.59)} & 55.24\textsubscript{(0.16)} & 65.36\textsubscript{(1.10)} \\
GCN & 52.19\textsubscript{(1.00)} & 61.01\textsubscript{(0.95)} & 46.07\textsubscript{(1.23)} & 56.61\textsubscript{(1.12)} & 42.37\textsubscript{(1.34)} & 53.07\textsubscript{(0.85)} & 16.20\textsubscript{(1.08)} & 50.73\textsubscript{(0.48)} & 65.21\textsubscript{(0.63)} & 68.76\textsubscript{(1.02)} \\
GAT & 54.25\textsubscript{(2.86)} & 63.55\textsubscript{(1.63)} & 46.58\textsubscript{(0.47)} & 60.72\textsubscript{(0.81)} & 37.55\textsubscript{(1.31)} & 52.63\textsubscript{(0.43)} & 18.61\textsubscript{(0.96)} & 51.83\textsubscript{(0.37)} & 59.33\textsubscript{(2.20)} & 65.94\textsubscript{(1.80)} \\
SAGE & 44.04\textsubscript{(1.28)} & 50.19\textsubscript{(0.89)} & 32.53\textsubscript{(2.83)} & 50.50\textsubscript{(1.70)} & 36.20\textsubscript{(0.84)} & 52.24\textsubscript{(0.19)} & 16.29\textsubscript{(0.46)} & 51.15\textsubscript{(0.41)} & 62.65\textsubscript{(0.42)} & \textbf{70.73}\textsubscript{(0.50)} \\
GT & 42.25\textsubscript{(1.81)} & 52.13\textsubscript{(2.09)} & 31.04\textsubscript{(2.55)} & 49.44\textsubscript{(2.11)} & 35.67\textsubscript{(0.87)} & 51.02\textsubscript{(0.17)} & 13.85\textsubscript{(0.83)} & 50.82\textsubscript{(0.54)} & 62.95\textsubscript{(1.19)} & 70.35\textsubscript{(1.05)} \\
DIR-GNN & \underline{\textbf{71.44}}\textsubscript{(0.96)} & 69.95\textsubscript{(1.33)} & 62.32\textsubscript{(0.72)} & \underline{\textbf{64.97}}\textsubscript{(0.63)} & 52.83\textsubscript{(7.05)} & 55.64\textsubscript{(3.03)} & 54.60\textsubscript{(2.34)} & 56.17\textsubscript{(1.02)} & 53.83\textsubscript{(4.77)} & 54.12\textsubscript{(3.56)} \\
GIB & 25.59\textsubscript{(18.05)} & 31.80\textsubscript{(16.51)} & 40.56\textsubscript{(16.48)} & 46.87\textsubscript{(13.35)} & 38.11\textsubscript{(5.95)} & 50.56\textsubscript{(0.69)} & 38.93\textsubscript{(8.60)} & 51.73\textsubscript{(2.61)} & 17.64\textsubscript{(8.56)} & 23.25\textsubscript{(8.04)} \\
VGIB & 20.64\textsubscript{(27.56)} & 26.54\textsubscript{(24.51)} & 54.25\textsubscript{(20.38)} & 56.74\textsubscript{(17.96)} & 39.00\textsubscript{(0.29)} & 50.06\textsubscript{(0.13)} & 36.58\textsubscript{(7.07)} & 50.60\textsubscript{(1.19)} & 45.45\textsubscript{(31.44)} & 47.58\textsubscript{(27.67)} \\
SEGNN & 56.47\textsubscript{(0.72)} & 59.51\textsubscript{(0.94)} & 60.23\textsubscript{(0.68)} & 63.10\textsubscript{(0.72)} & 54.32\textsubscript{(0.52)} & 54.61\textsubscript{(0.75)} & 55.81\textsubscript{(1.45)} & \underline{\textbf{56.36}}\textsubscript{(1.57)} & 35.41\textsubscript{(0.52)} & 39.84\textsubscript{(0.52)} \\
\model{} & 70.48\textsubscript{(2.31)} & \underline{\textbf{70.80}}\textsubscript{(1.28)} & \underline{\textbf{63.39}}\textsubscript{(0.37)} & 63.76\textsubscript{(0.38)} & \underline{\textbf{56.95}}\textsubscript{(0.18)} & \underline{\textbf{56.91}}\textsubscript{(0.19)} & \underline{\textbf{55.02}}\textsubscript{(0.67)} & 55.12\textsubscript{(0.68)} & \underline{\textbf{69.17}}\textsubscript{(0.45)} & \underline{70.45}\textsubscript{(0.51)} \\
\bottomrule
\end{tabular}

\end{table}

\end{document}